\documentclass[10pt,journal,compsoc]{IEEEtran}
\usepackage{algorithm}
\usepackage{algpseudocode}  
\usepackage{amsmath}  
\usepackage{multirow, makecell}
\usepackage{booktabs}
\usepackage{graphicx}
\usepackage{color}

\usepackage{svg}

\usepackage{subfigure}
\usepackage{pifont}

\usepackage{amssymb}
\usepackage{amsthm}
\ifCLASSOPTIONcompsoc
  \usepackage[nocompress]{cite}
\else
  \usepackage{cite}
\fi

\ifCLASSINFOpdf
\else
\fi

\begin{document}
%
\title{Shared Growth of Graph Neural Networks via Prompted Free-direction Knowledge Distillation}
%
%
%
%
\author{
Kaituo~Feng,
Yikun~Miao,
Changsheng~Li,
Ye~Yuan,
Guoren~Wang

\IEEEcompsocitemizethanks{
\IEEEcompsocthanksitem 
Kaituo Feng,  Yikun Miao, Changsheng Li, Ye Yuan, and Guoren Wang are with the School of Computer Science and Technology, Beijing Institute of Technology, China. E-mail: \{kaituofeng@gmail.com; 1120201064@bit.edu.cn; lcs@bit.edu.cn; yuan-ye@bit.edu.cn; wanggrbit@126.com.\}
\IEEEcompsocthanksitem Corresponding author: Changsheng Li.
}}

\markboth{Submitted to IEEE Transactions on Pattern Analysis and Machine Intelligence}%
{Shell \MakeLowercase{\textit{et al.}}: Bare Demo of IEEEtran.cls for Computer Society Journals}
%



\IEEEtitleabstractindextext{%
\begin{abstract}
Knowledge distillation (KD) has shown to be effective to boost the performance of graph neural networks (GNNs), where the typical objective is to distill knowledge from a deeper teacher GNN into a shallower student GNN.
However, it is often quite challenging to train a satisfactory deeper GNN due to the well-known over-parametrized and over-smoothing issues, leading to invalid knowledge transfer in practical applications. 
In this paper, we propose the first \textbf{Free}-direction \textbf{K}nowledge \textbf{D}istillation framework via  reinforcement learning for GNNs, called \textbf{FreeKD}, which is no longer required to provide a deeper well-optimized teacher GNN. Our core idea is to collaboratively learn two shallower GNNs in an effort to exchange knowledge between them via reinforcement learning in a hierarchical way.
As we observe that one typical GNN model often exhibits better and worse performances at different nodes during training, we devise a dynamic and free-direction knowledge transfer strategy that involves two levels of actions: 
1) node-level action determines the directions of knowledge transfer between the corresponding nodes of two networks; and then 2) structure-level action determines which of the local structures generated by the node-level actions to be propagated.
Additionally, considering that different augmented graphs can potentially capture distinct perspectives or representations of the graph data, we propose FreeKD-Prompt that learns undistorted and diverse augmentations based on prompt learning for exchanging varied knowledge.
Furthermore, instead of confining knowledge exchange within two GNNs, we develop FreeKD++ and FreeKD-Prompt++ to enable free-direction knowledge transfer among multiple shallow GNNs.
Extensive experiments on five benchmark datasets demonstrate our approaches outperform the base GNNs in a large margin, and shows their efficacy to various GNNs.
More surprisingly, our FreeKD has comparable or even better performance than traditional KD algorithms that distill knowledge from a deeper and stronger teacher GNN.
\end{abstract}

}

\maketitle

\IEEEdisplaynontitleabstractindextext

%
\IEEEpeerreviewmaketitle

\section{Introduction}

\IEEEPARstart{G}{raph} data has witnessed a surge in prevalence and ubiquity due to the rapid development of the Internet. This includes diverse domains such as social networks \cite{hamilton2017inductive} and citation networks \cite{sen2008collective}. In order to effectively handle the inherent complexities of graph-structured data, graph neural networks (GNNs) have emerged as a powerful approach for learning node embeddings by aggregating feature information from neighboring nodes \cite{velivckovic2017graph}.
Over the past decade, the research community has proposed various graph neural networks, driven by their remarkable ability to model intricate data relationships \cite{hamilton2017inductive,velivckovic2017graph,bianchi2021graph,isufi2021edgenets, bessadok2022graph,xie2022self}. The representative works include GraphSAGE \cite{hamilton2017inductive}, GAT \cite{velivckovic2017graph}, GCN \cite{kipf2016semi}, NLGCN \cite{liu2021non}, LGLP \cite{cai2021line}, GSN \cite{bouritsas2022improving} etc.


Recently, there has been a growing interest in extending the concept of knowledge distillation (KD) to graph neural networks (GNNs) as a means to further enhance their performance \cite{yang2020distilling,yang2021extract,deng2021graph}. These approaches aim to optimize shallower student GNN models by distilling knowledge from deeper teacher GNN models. For example, LSP \cite{yang2020distilling} introduces a local structure preserving module to transfer topological structure information from a  teacher GNN model. 
The work in \cite{yan2020tinygnn} introduces a neighbor knowledge distillation approach that aims to bridge the neighbor information gap between a shallower student GNN model and a deeper teacher GNN model.
Furthermore, GFKD \cite{deng2021graph} devises a data-free knowledge distillation strategy for GNNs, enabling the transfer of knowledge from a teacher GNN model through the generation of fake graphs.


The aforementioned methods follow the same teacher-student architecture commonly used in traditional knowledge distillation techniques \cite{bucilu2006model,hinton2015distilling}, which resort to a deeper well-optimized teacher GNN for distilling knowledge.  However, when applying such an architecture to GNNs, it often encounters the following limitations: first, training a satisfactory teacher GNN is often difficult and inefficient. 
As we know, the performance of deeper GNN models is often compromised by the existing over-parameterized and  over-smoothing issues. Additionally, training a deeper, well-optimized model typically requires a large amount of data and entails high computational costs. Second, according to \cite{yan2020tinygnn,yuan2021reinforced}, a stronger teacher model may not necessarily result in a better student model. The mismatch in representation capacities between teacher and student models can make it challenging for the student model to mimic the outputs of a too strong teacher model. 
As a result, finding an optimal teacher GNN for a student GNN becomes a challenge in practical applications.
Given the multitude of powerful GNN models that have emerged in the last decade \cite{wu2020comprehensive}, this gives rise to one intuitive thought:  \textbf{Can we explore a new knowledge distillation architecture to enhance the performance of GNNs while bypassing the challenges associated with training a deeper, well-optimized teacher GNN?}

\begin{figure}
  \centering
  \includegraphics[width=1.0\linewidth]{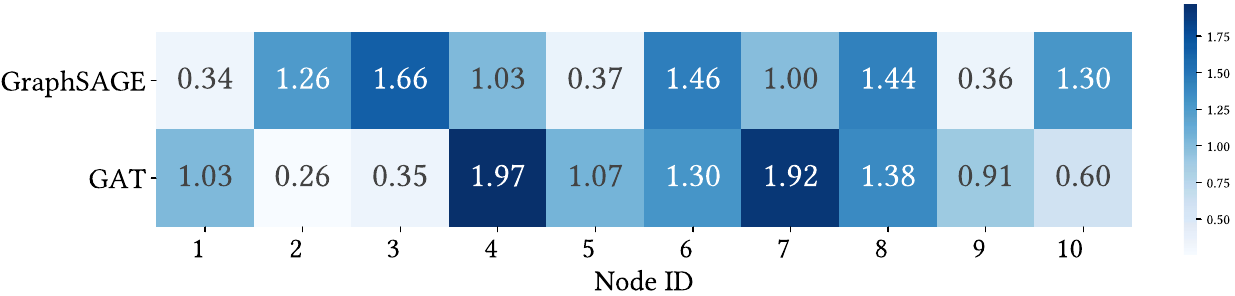}
  \vspace{-0.2in}
  \caption {Cross entropy losses for nodes with ID  from $1$ to $10$ on the Cora dataset obtained by two typical GNN models, GraphSAGE \cite{hamilton2017inductive} and GAT \cite{velivckovic2017graph}, after training 20 epochs. The value in each block denotes the corresponding loss.}
  \label{cross}
  \vspace{-0.2in}
\end{figure}


In light of these, we propose a new knowledge distillation framework, \textbf{Free}-direction \textbf{K}nowledge \textbf{D}istillation based on Reinforcement learning tailored for GNNs, called \textbf{FreeKD}. Instead of relying on a deeper well-optimized teacher GNN for unidirectional knowledge transfer, we collaboratively learn two  shallower GNNs in an effort to distill knowledge from each other using reinforcement learning in a hierarchical manner. This idea stems from our observation that  GNN models often exhibit varying performances across different nodes during training. For instance, as illustrated in Fig. \ref{cross}, GraphSAGE \cite{hamilton2017inductive} has lower cross-entropy losses at nodes with ID$=\{1,4,5,7,9\}$, while GAT \cite{velivckovic2017graph} performs better at the remaining nodes.

Based on this observation, we explore to design a free-direction knowledge distillation strategy to dynamically exchange useful knowledge between two shallower GNNs  to benefit from each other.
Considering that the direction of distilling knowledge for each node will have influence on the other nodes, we thus regard determining the directions for different nodes as a sequential decision making problem. Meanwhile, since the selection of the directions is a discrete problem, we can not optimize it by stochastic gradient descent based methods \cite{wang2019minimax}. Thus, we address this problem via reinforcement learning in a hierarchical way. Our hierarchical reinforcement learning algorithm consists of  two levels of actions: Level 1, called  node-level action, is used to distinguish which GNN is chosen to distill knowledge to the other GNN for each node.
After determining the direction of knowledge transfer for each node, we expect to propagate not only the soft label of the node, but also its neighborhood relations.
Thus level 2, called  structure-level action, decides which of the local structures generated by our node-level actions to be propagated.
One may argue that we could directly use the loss, e.g., cross entropy, to decide the directions of  node-level knowledge distillation.
However, this heuristic strategy only considers the performance of the node itself, but neglects its influence on other nodes, thus might lead to a sub-optimal solution.
Our experimental results also verify our reinforcement learning based strategy significantly outperforms the above heuristic one.
By leveraging FreeKD, the two shallower GNNs can engage in a flexible and dynamic exchange of knowledge, leading to a mutual improvement.

In addition, considering that different augmented graph views may capture distinct perspectives or representations of the graph data \cite{miao2023triplet,ko2023signed,zhang2023graph,hassani2020contrastive}, we intend to adopt graph augmentation methods (e.g. DropEdge \cite{rong2019dropedge}, DropNode \cite{feng2020graph}) to generate different augmented views of the input graph for distilling diverse knowledge from different augmented perspectives. 
However, this straightforward scheme could encounter the following two limitations: (1) The widely-used graph augmentation methods might cause the distortion of the original graph without carefully calibrations, resulting in unexpected semantic changes. 
For instance, in biological graph, if a carbon atom is dropped by DropNode from the phenyl ring in an aspirin, the aromatic system will be destroyed and the aspirin becomes a alkene chain \cite{lee2022augmentation}. As for social networks, randomly dropping edges between connected nodes (e.g. dropping the link between two hub nodes) could also introduce unexpected semantic changes. (2) Due to the randomness for augmentation operations, it's difficult to ensure that the multiple views obtained by different graph augmentations can lead to diverse knowledge for distillation. 
For example, randomly dropping edges or nodes can sometimes result in generating similar or redundant views of the original graph.
This lack of control over the diversity of the generated views might hinder the two GNNs from distilling varying knowledge across multiple augmentation views.

To tackle with the above limitations, we propose a new method FreeKD-Prompt based on prompt learning.
FreeKD-Prompt aims to derive augmentation views in an adaptive manner, allowing for the distillation of undistorted and diverse knowledge.
This inspiration draws from a recent work \cite{sun2023all}, which unifies the format of the prompt in the language area and the graph area and utilizes the proposed graph prompt for effective multi-task learning. 
Different from this work, our approach constitutes the first attempt to adopt prompt learning to learn graph augmentations for distillation. 
Our key idea is to learn a group of prompt graphs and insert them to the original input graph for augmentations. 
The optimization of prompt graphs is guided by two distinctive loss functions. Firstly, an information preservation loss is devised through mutual information maximization, encouraging the augmented graph to retain the semantic information of the original graph. Secondly, a diversity loss is formulated to enhance the diversity of the augmented graphs, facilitating the distillation of varied knowledge.
By optimizing these two losses, the resulting augmented graphs can effectively mitigate drastic or unexpected semantic changes while still maintaining diversity to encompass various knowledge. Consequently, FreeKD-Prompt facilitates a more comprehensive knowledge exchange between the two GNNs.

Furthermore, in contrast to confining knowledge transfer within two GNNs, our approach goes a step further by developing FreeKD++ and FreeKD-Prompt++ to facilitate free-direction knowledge transfer among multiple GNNs.
In these schemes, each pair of GNNs among the multiple GNNs exchanges knowledge in a free-direction manner.
This comprehensive knowledge exchange enables to share valuable knowledge among  multiple GNNs, empowering them to mutually enhance their performance.
    
Our main contributions can be summarized as:
\begin{itemize}
\item We propose a new knowledge distillation architecture to mutually distill knowledge from two shallower GNN models, avoiding requiring a deeper well-optimized teacher model for distilling knowledge. The proposed framework is general and principled, which can be naturally compatible with GNNs of different architectures. 

\item We devise a  free-direction knowledge distillation strategy via a hierarchical reinforcement learning  algorithm, which can dynamically manage the directions of knowledge transfer between two GNNs from both node-level and structure-level aspects. In addition, We develop FreeKD++ to enable free-direction knowledge distillation among  multiple GNNs. 

\item 
Furthermore, we present FreeKD-Prompt as a means to facilitate the exchange of diverse knowledge between GNNs utilizing multiple prompt-based graph augmentation inputs. To the best of our knowledge, we are the first to utilize prompt learning to learn undistorted and diverse graph augmentations.
 
\item Extensive experiments on five benchmark datasets demonstrate our proposed approaches promote the performance of the shallower GNNs in a large margin, and are valid to various GNNs. More surprisingly,  our proposed methods are comparable to or even better than traditional KD algorithms distilling knowledge from a deeper and stronger teacher GNN.
\end{itemize}

Note that this article is a substantial extension of our previous \emph{KDD} work \cite{feng2022freekd}.
Compared to the conference version, we have made significant extensions in this journal manuscript: 
(1) We explore learning undistorted and diverse graph augmentations based on prompt learning and propose a new method FreeKD-Prompt,  with the goal to utilize the derived augmentations for exchanging diverse knowledge between GNNs. 
(2) In contrast to restricting knowledge transfer between two GNNs, we introduce FreeKD++ to facilitate knowledge exchange among multiple GNNs in a free-direction manner, fostering a more flexible and open approach. 
(3) We update extensive new experiments to further demonstrate the effectiveness of our proposed methods. For example, we incorporate two experiments aimed at gaining deeper insights into the reasons behind the effectiveness of our FreeKD. 
(4) We include a more comprehensive survey on the related works about the graph neurual networks, knowledge distilltions for GNNs, reinforcement learning and prompt learning. 

The rest of the paper is organized as follows. In Section 2, we review the related works. Section 3 provides the preliminaries  and introduces our proposed method in detail. Then, we conduct extensive experiments to validate the effectiveness of our proposed methods in Section 4. Finally, Section 5 concludes this paper.

\section{Related Work}
This work is related to graph neural networks, graph-based knowledge distillation, and reinforcement learning. 
Thus, we will briefly introduce them, respectively.

\subsection{Graph Neural Networks}

Graph neural networks have demonstrated promising results in processing graph data, whose basic goal is to learn node embeddings by aggregating nodes' neighbor information \cite{kipf2016semi,hamilton2017inductive}.
In recent years, there has been a surge of proposed GNN models \cite{kipf2016semi,li2022robust,feng2023towards,bianchi2021graph,isufi2021edgenets,velivckovic2017graph}. For instance, GCN \cite{kipf2016semi} introduces a convolutional neural network architecture designed for graph data.
GraphSAGE \cite{hamilton2017inductive} proposes an efficient sample strategy  to aggregate neighbor nodes, and can perform inductive learning in large-scale graphs with limited computational resources. 
GAT \cite{velivckovic2017graph} introduces a self-attention mechanism to GNNs, enabling the assignment of varying weights to different neighbors based on their importance. 
 \cite{xu2018powerful} demonstrates that the upper bound  of the representation ability of GNN that is the Weisfeiler-Lehman isomorphism test \cite{weisfeiler1968reduction}, and builds a GNN model that could reach to this upper bound. 
SGC \cite{wu2019simplifying} proposes a simplified version of GCN by removing nonlinearities and weight matrices between consecutive convolutional layers. This simplification not only reduces the computational complexity but also maintains competitive performance, making it a more efficient choice for certain graph-based applications.
ROD \cite{zhang2021rod} proposes an ensemble learning based GNN model to fuse knowledge in multiple hops, which facilitates a comprehensive utilization of knowledge across different hops.
APPNP \cite{gasteiger2018predict} analyzes the relationship between GCN and PageRank \cite{page1999pagerank}, and proposes a propagation model combined with a personalized PageRank.
GCNII \cite{chen2020simple} attempts to mitigate the problem of over-smoothing in deep graph neural networks by introducing the initial residual and identity mapping techniques. 
The initial residual technique establishes a skip connection from the input layer, while the identity mapping technique incorporates an identity matrix into the weight matrix at each layer.
Graphormer \cite{ying2021transformers} builds a graph transformer architecture and designs several structural encoding strategies for better capturing the topological information in graph.
Being orthogonal to the above approaches developing different powerful GNN models, we concentrate on developing a new knowledge distillation framework on the basis of various GNNs.

\subsection{Knowledge Distillation for GNNs}

Knowledge distillation (KD) has been widely studied in computer vision \cite{chen2017learning,liu2018multi}, natural language processing \cite{kim2016sequence,liu2019improving}, etc. 
The fundamental idea behind KD is to transfer knowledge from a larger teacher model to a smaller student model, so as to improve the performance of the student model \cite{hinton2015distilling}.
Recently, a few KD methods have proposed for GNNs \cite{yang2020distilling,guo2022boosting, feng2022freekd, yang2021extract,joshi2022representation}. For instance, 
LSP \cite{yang2020distilling} transfers the topological structure knowledge from a pre-trained deeper teacher GNN to a shallower student GNN.
CPF \cite{yang2021extract} designs a student architecture that is a combination of a parameterized label propagation and MLP layers.
GFKD \cite{deng2021graph} proposes a data-free knowledge distillation method for GNNs, by leveraging generated fake graphs to transfer knowledge from a teacher GNN model. 
The authors in \cite{yan2020tinygnn} propose a neighbor knowledge distillation method for GNNs to address the information gap in neighbor representation between a shallower student GNN model and a deeper teacher GNN model. 
The work in \cite{chen2020self} studies a self-distillation framework, and proposes an adaptive discrepancy retaining regularization to transfer knowledge.
GNN-SD \cite{chen2020self} proposes to distill knowledge from the shallow layers to the deep layers in one GNN model.
RDD \cite{zhang2020reliable} is a semi-supervised knowledge distillation method for GNNs. It online learns a complicated teacher GNN model by ensemble learning, and distills  knowledge from a generated teacher model into a student model. 
Different from them, we focus on studying a new free-direction knowledge distillation architecture, with the purpose of dynamically exchanging knowledge between shallower GNNs. 
    

\subsection{Reinforcement Learning}

Reinforcement learning aims at training agents to make optimal decisions by learning from interactions with the environment \cite{arulkumaran2017deep, chen2018recurrent}. Reinforcement learning mainly has two genres \cite{arulkumaran2017deep}: value-based methods and policy-based methods. 
Value-based methods, such as those based on deep Q-networks (DQN) \cite{mnih2015human}, estimate the expected reward corresponding to different actions. These methods learn the value functions, which quantify the expected rewards to select the optimal actions \cite{van2016deep}. On the other hand, policy-based methods, like the REINFORCE algorithm \cite{williams1992simple}, directly determine actions based on the output probabilities generated by the agent's policy. Instead of explicitly estimating the value function, policy-based methods focus on improving the policy itself to maximize the cumulative rewards \cite{huang2022reward}.
An interesting hybrid approach that combines these two genres is the actor-critic architecture \cite{haarnoja2018soft, luo2019end}. 
The actor-critic architecture utilizes the value-based method as a critic to estimate the expected reward, and employs the policy-based method as an actor to take actions \cite{haarnoja2018soft}. 
The critic's value estimations guide the actor's decision-making process, providing a measure of the expected reward. Meanwhile, the actor explores different actions based on the learned policy to adapt its behavior and improve performance. 
Until now,  reinforcement learning has been taken as a popular tool to solve various tasks, such as recommendation systems \cite{guo2022reinforcement}, 
anomaly detection \cite{ma2021comprehensive}, 
autonomous driving \cite{shalev2016safe}, etc.
In this paper, we explore  reinforcement learning for graph data based knowledge distillation. 

\subsection{Prompt Learning}
Prompt learning is an emerging paradigm in natural language processing (NLP) that typically leverages prompts or pre-defined instructions (e.g. task descriptions) to guide the model's behavior in generating text outputs \cite{liu2023pre,brown2020language}. 
This paradigm has been widely used in NLP for the fast adaptation of pre-trained language models on downstream tasks \cite{brown2020language,shin2020autoprompt,li2021prefix,jiang2020can,liu2021p}. For example, GPT-3 \cite{brown2020language} utilizes hand-crafted prompts to guide its text generation, enabling the adaptation to a variety of tasks. Instead of designing the prompt manually, AutoPrompt \cite{shin2020autoprompt} proposes an automated method based on gradient-guided search to derive prompts for a variety of tasks. In contrast to the discrete prompts, Prefix-tuning \cite{li2021prefix} learns continuous prompt vectors in the prefix of the transformer token sequence for adapting to different downstream tasks. 
While prompt learning is a well-explored concept in NLP, it remains highly nascent in the field of graph learning, with only a mere handful of works dedicated to this emerging research direction \cite{sun2022gppt,sun2023all,zhu2023sgl}. For instance, GPPT \cite{sun2022gppt} pre-trains the GNN model on an edge prediction task and reformulates the downstream node classification task as edge prediction based on a graph prompt function. ProG \cite{sun2023all} introduces a prompt graph design that unifies the format of the prompt in the NLP field and the graph field for effective multi-task learning and adaptation. Different from these works, we explore learning graph prompts to generate undistorted and diverse graph augmentations for distilling varied knowledge.

\section{The Proposed Method}
\begin{figure*}[htbp]
  \centering
  \includegraphics[width=0.95\linewidth]{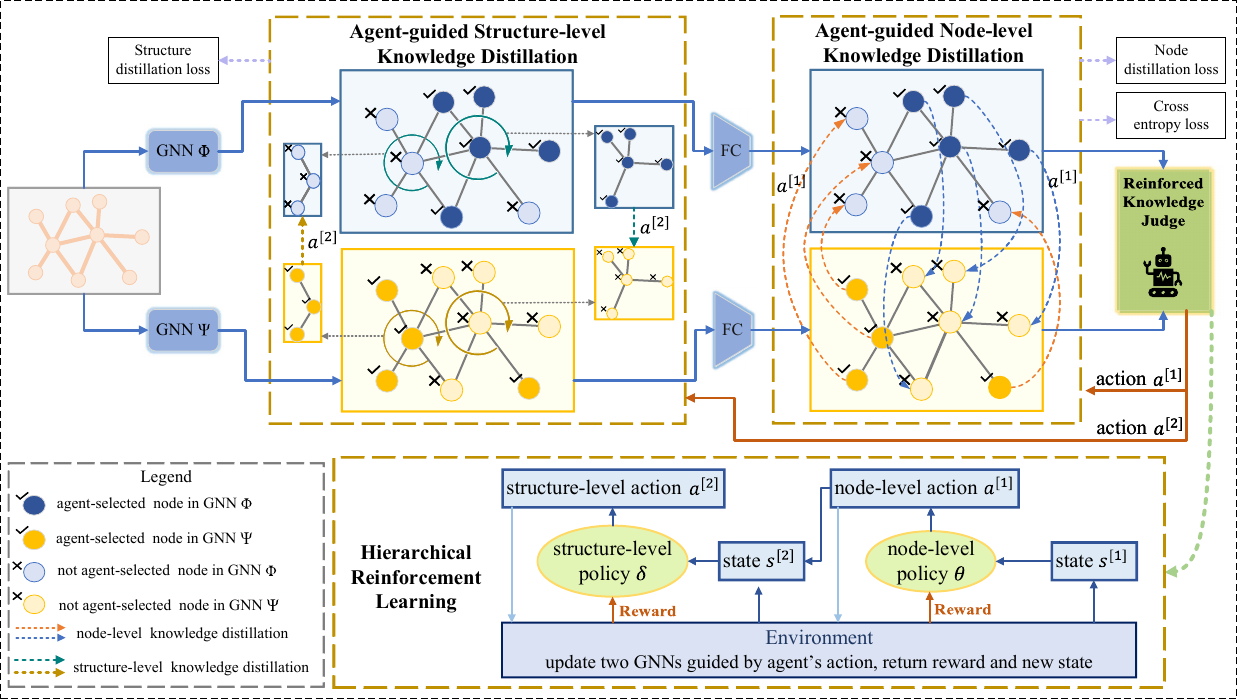}
    \vspace{-0.1in}
  \caption{
    An illustration of the FreeKD framework. FreeKD can manage the knowledge distillation directions between two GNN models via a  hierarchical reinforcement learning that contains two-level actions. The first level of actions are designed to determine the distillation direction for each node, in order to propagate the node's soft label. And then the second level of actions are used to decide which of   the local structures generated based on node-level actions to be propagated.
  }
  \vspace{-0.15in}
  \label{main}
\end{figure*}

    In this section, we first elaborate the details of our FreeKD framework that is shown in Fig. \ref{main}, and then introduce our FreeKD++. 
    Before introducing them, we first give some notations and  preliminaries.

\subsection{Preliminaries}
	Let $\mathbf{G}=(\mathbf{V},\mathbf{E},\mathbf{X})$ denote a graph, where $\mathbf{V}$ is the set of nodes and $\mathbf{E}$ is the set of edge. $\ \mathbf{X}\in\mathbb{R}^{N\times d}$ is the feature matrix of nodes, where $N$ is the number of nodes and $d$ is the dimension of node features.  Let $\mathbf{x}_i$ be the feature representation of node $i$ and $y_i$ be its class label.
	The neighborhood set of node $i$ is ${\mathcal{N}(i)}=\{\ j\in \mathbf{V}\ |\ (i,j)\in \mathbf{E}\ \}$.
	Currently, graph neural networks (GNNs) have become one of the most popular models for handling graph data.
	 GNNs can learn the embedding $\mathbf{h}_i^{(l)}$  for node $i$ at the $l$-th layer  by the following formula:
\begin{align}
  \mathbf{h}_i^{(l)}\!\!=\!AGGREGATE(\mathbf{h}_i^{(l-1)}\!\!\!,\{\mathbf{h}_j^{(l-1)}|\ \!j\!\!\in\!{\mathcal{N}(i)}\},\!{\ \!\!\mathbf{W}^{(l)}}),
\end{align}
where $AGGREGATE$ is an aggregation function, and it can be defined in many forms, e.g., mean aggregator \cite{hamilton2017inductive}.  $\mathbf{W}^{(l)}$ is the learnt parameters in the  $l$-th layer of the network. The initial feature of each node $i$ can be used as the input of the first layer, i.e., $\mathbf{h}_i^{(0)}=\mathbf{x}_i$.




Being orthogonal to those works developing various GNN models, our goal is to explore a new knowledge distillation framework for promoting the performance of GNNs, while addressing the issue involved because of producing a deeper teacher GNN model in the existing KD methods. 

\subsection{Overview of Framework}
As shown in Fig. \ref{cross}, we observe typical GNN models often have different performances at different nodes during training.
Based on this observation, we intend to dynamically exchange useful knowledge between two shallower GNNs, so as to benefit from each other. 
However, a challenging problem is attendant upon that: how to decide the directions of knowledge distillation for different nodes during training. 
To address this, we propose to manage the directions of knowledge distillation via reinforcement learning, where we regard the directions of knowledge transfer for different nodes as a sequential decision making problem \cite{pednault2002sequential}. Consequently, we propose a free-direction knowledge distillation framework via a hierarchical reinforcement learning, as shown in Fig. \ref{main}.
In our framework, the hierarchical reinforcement learning can be taken as a reinforced knowledge judge that consists of two levels of actions: 1) Level 1, called node-level action, is used to decide the distillation direction of each node for propagating the soft label; 2) Level 2, called structure-level action, is used to determine which of  the local structures generated via node-level actions to be propagated.

 Specifically, the reinforced knowledge judge (we call it agent for convenience) interacts with the environment constructed by two GNN models in each iteration, as in Fig. \ref{main}.
	It receives the soft labels and cross entropy losses for a batch of nodes, and regards them as its node-level states. 
	The agent then samples sequential node-level actions for nodes according to a learned policy network, where each action decides the direction of knowledge distillation for propagating node-level knowledge. 
	Then, the agent receives the structure-level states and produces structure-level actions to decide which of the local structures generated on the basis of node-level actions to be propagated.
	After that, the two GNN models are trained based to the agent's actions with a new loss function.
	Finally, the agent calculates the reward for each action to train the policy network, where the agent's target is to maximize the expected reward.
	This process is repeatedly iterated until convergence.

	We first  give some notations for convenient presentation, before introducing how to distill both node-level and structure-level knowledge.
	Let $\Phi$ and $\Psi$ denote two GNN models (Note that both models have a small number of layers.), respectively.  $\mathbf{h}_i^\Phi$ and $\mathbf{h}_i^\Psi$ denote the learnt representations of node $i$ obtained by $\Phi$ and $\Psi$, respectively.
	Let $\mathbf{p}_i^\Phi$ and $\mathbf{p}_i^\Psi$ be the predicted probabilities of the two GNN models for node $i$ respectively. We regard them as the soft labels.
	In addition,  $L_{CE}^\Phi(i)$ and $L_{CE}^\Psi(i)$ denote the cross entropy losses of node $i$ in $\Phi$ and $\Psi$, respectively.

 \vspace{-0.1in}

\subsection{Agent-guided Node-level Knowledge Distillation}
In this section, we introduce our reinforcement learning based strategy  to dynamically distill node-level knowledge  between two GNNs, $\Phi$ and $\Psi$. 

\subsubsection{Node-level State}  We concatenate the following features as the node-level state vector $s_i^{[1]}$ for node $i$: 

(1) Soft label vector of node $i$ in GNN $\Phi$.

(2) Cross entropy loss of node $i$ in GNN $\Phi$.

(3) Soft label vector of node $i$ in GNN $\Psi$.

(4) Cross entropy loss of node $i$ in GNN $\Psi$.

	The first two kinds of features are based on the intuition that the cross entropy loss and soft label can quantify the useful knowledge for node $i$ in GNN $\Phi$ to some extent. 
	The last two kinds of features have the same function for $\Psi$.
	Since these features can measure the knowledge each node contains to some extent, we use them as the feature of the node-level state for predicting the node-level actions.
 
	Formally, the state $s_i^{[1]}$ for the node $i$ is expressed as:
\begin{equation}\label{n-state}
    \mathbf{s}_i^{[1]}=CONCAT(\mathbf{p}_i^\Phi,L_{CE}^\Phi(i),\mathbf{p}_i^\Psi,L_{CE}^\Psi(i)),
\end{equation}
where $CONCAT$ is the concatenation operation.

\subsubsection{Node-level Action} 
	The node-level action $a_i^{[1]}\in\{0,1\}$ decides the direction of knowledge distillation for node $i$.
	$a_i^{[1]}=0$ means transferring knowledge from GNN $\Phi$ to GNN $\Psi$ at node $i$, while $a_i^{[1]}=1$ means the distillation direction from $\Psi$ to  $\Phi$. 
	If $a_i^{[1]}=0$, we define node $i$ in $\Phi$ as \emph{agent-selected node}, otherwise,  we define node $i$ in $\Psi$ as \emph{agent-selected node}.
	The actions are sampled from the probability distributions produced by a node-level policy function $\pi_\mathbf{\theta}$, where $\mathbf{\theta}$ is the trainable parameters in the policy network and $\pi_\mathbf{\theta}\left(s_i^{[1]},a_i^{[1]} \right)$ means the probability to take action $a_i^{[1]}$ over the state $s_i^{[1]}$. In this paper, we adopt a three-layer MLP with the $tanh$ activation function as our node-level policy network.

\begin{figure}
\centering
\subfigure[Node-level distillation]{
\begin{minipage}[t]{0.49\linewidth}
\centering
\includegraphics[width=1.57in]{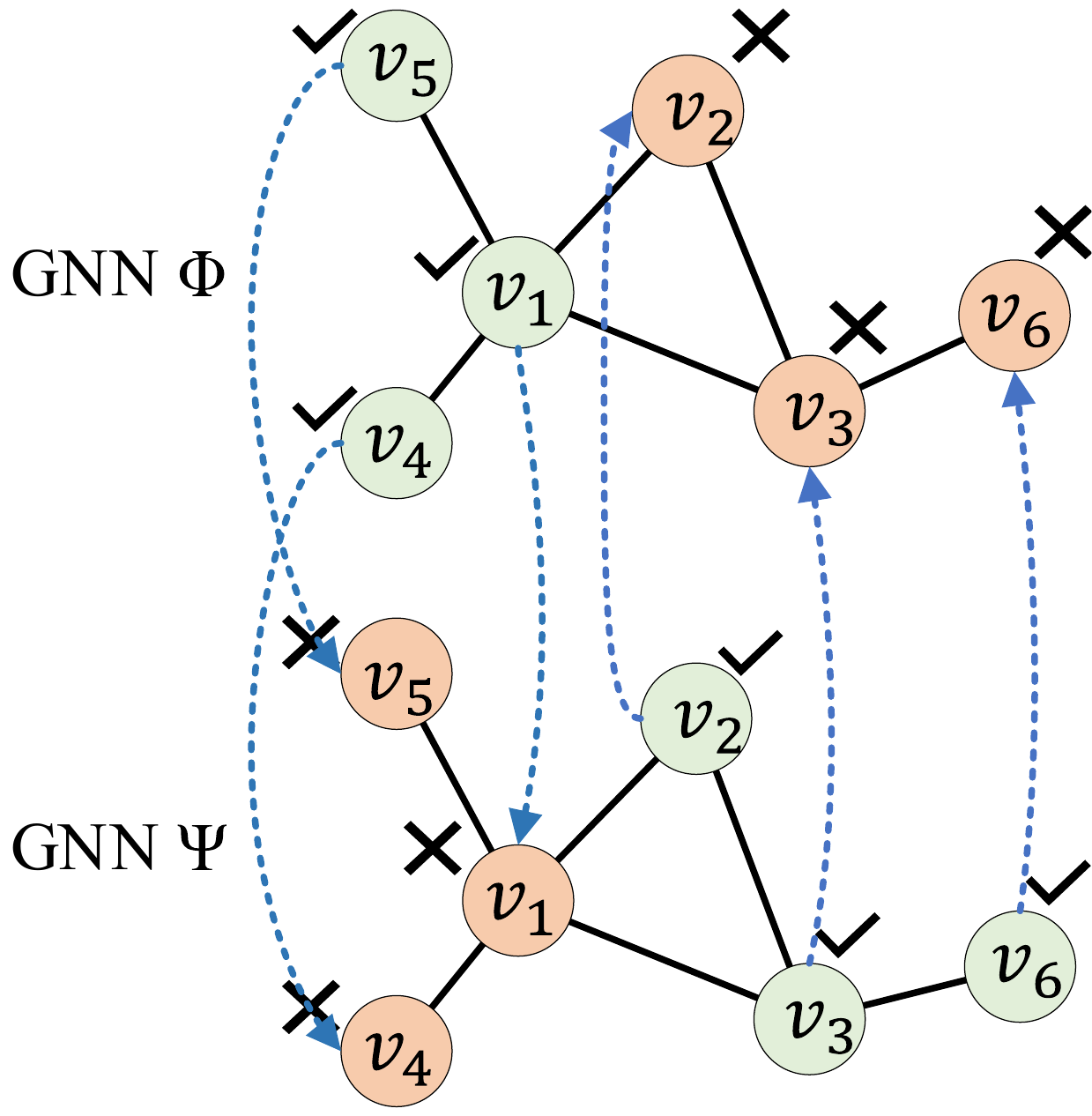}
\end{minipage}%
}%
\subfigure[Structure-level distillation]{
\begin{minipage}[t]{0.49\linewidth}
\centering
\includegraphics[width=1.3in]{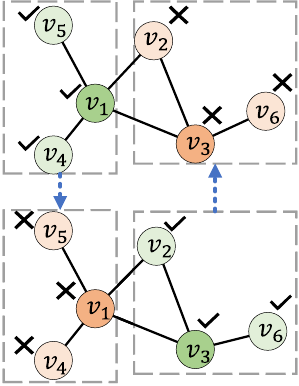}
\end{minipage}%
}%
\centering
\vspace{-0.1in}
\caption{Illustration of both node-level and structure-level knowledge distillation.}
\vspace{-0.2in}
\label{transfer}
\end{figure}

\subsubsection{Node-level Knowledge Distillation} 
    After determining the direction of knowledge distillation for each node, the two GNN models, $\Phi$ and $\Psi$, can exchange beneficial node-level knowledge. We take Fig. \ref{transfer}(a) as an example to illustrate our idea. In Fig. \ref{transfer}(a), the agent-selected nodes $\{v_1, v_4, v_5\}$ in GNN $\Phi$ will serve as the distilled nodes to transfer knowledge to the nodes  $\{v_1, v_4, v_5\}$  in GNN $\Psi$. In the meantime, the agent-selected nodes $\{v_2, v_3, v_6\}$ in $\Psi$ will be used as the distilled nodes to distill knowledge for the nodes $\{v_2, v_3, v_6\}$ in $\Phi$. 
    In order to transfer node-level knowledge, we utilize the KL divergence to measure the distance between the soft labels of the same node in the two GNN models, and  propose to minimize a new loss function for each GNN model as:  
\begin{align}
\setlength{\abovedisplayskip}{0.05in}
\setlength{\belowdisplayskip}{0.05in}
\mathcal{L}_{node}^\Psi&=\sum_{i=1}^{N}{(1-a_i^{[1]}) KL(\mathbf{p}_i^\Phi||\mathbf{p}_i^\Psi)}\label{ce1}\\
\mathcal{L}_{node}^\Phi&=\sum_{i=1}^{N}{a_i^{[1]}KL(\mathbf{p}_i^\Psi||\mathbf{p}_i^\Phi)}\label{ce2},
\end{align}
where the value of $a_i^{[1]}$ is 0 or 1. 
When $a_i^{[1]}=0$, we use the $KL$ divergence to make the probability distribution $\mathbf{p}_i^\Psi$ match $\mathbf{p}_i^\Phi$ as much as possible, enabling the knowledge from $\Phi$ to be transferred to $\Psi$ at node $i$, and vice versa for $a_i^{[1]}=1$.
Thus, by minimizing the two loss functions $\mathcal{L}_{node}^\Phi$ and $\mathcal{L}_{node}^\Psi$, we can reach the goal of dynamically exchanging useful node-level knowledge between two GNN models, and thus obtaining gains from each other. 

\subsection{Agent-guided Structure-level Knowledge Distillation}

As we know, the structure information is important for graph learning \cite{yang2020distilling}. Thus, we attempt to dynamically transfer structure-level knowledge between $\Phi$ and $\Psi$.
It is worth noting that we don't propagate all neighborhood information of one node as structure-level knowledge. Instead, we only propagate a neighborhood subset of the node, which is comprised of agent-selected nodes. This is because we think agent-selected nodes contain more useful knowledge. 
We take Fig. \ref{transfer}(b) as an example to illustrate it. $v_1$ is an agent-selected node in $\Phi$. When transferring its local structure information to $\Psi$, we only transfer the local structure composed of $\{v_1, v_4, v_5\}$.
In other words, the local structure of node $v_1$ we consider to transfer is made up of agent-selected nodes.  We call it agent-selected neighborhood set.
Moreover, considering the knowledge of the local structure in graphs is not always reliable \cite{zhang2020reliable,chen2021topology}, we design a reinforcement learning based strategy to distinguish which of the local structures to be propagated.
Next, we introduce it in detail.

\subsubsection{Structure-level State} 
We adopt the following features as the structure-level state vector $s_i^{[2]}$ for the local structure of node $i$:

(1) Node-level state of node $i$.

(2) Center similarity of node $i$'s agent-selected neighborhood set in the distilled network. 

(3) Center similarity of the same node set as node $i$'s agent-selected neighborhood set in the guided network. 

Since the node-level state contains much information related to the local structures, we use the node-level state as the first feature of structure-level state. 
As \cite{xie2020gnns} points out, the center similarity can indicate the performance of GNNs, where the center similarity measures the degree of similarity between the node and its neighbors. In other words, if center similarity  is high, the structure information should be more reliable. Thus, we also take the center similarity as another feature.
Motivated by \cite{xie2020gnns}, we present a similar strategy to calculate the center similarity as:

    First, let $\mathbf{M}_i^\Phi$ and $\mathbf{M}_i^\Psi$ denote the  agent-selected neighborhood set of node $i$ in $\Phi$ and $\Psi$, respectively. Formally, 
    \begin{align}
        &\mathbf{M}_i^\Phi=\{v\ |\ a_i^{[1]}=0\ , a_v^{[1]}=0, and\ \left(i,v\right)\in \mathbf{E}\}\\
        &\mathbf{M}_i^\Psi=\{v\ |\ a_i^{[1]}=1\ , a_v^{[1]}=1, and\ \left(i,v\right)\in \mathbf{E}\}.
    \end{align}
Then, we calculate the center similarity as:
    \begin{equation} \nonumber
    \mathbf{u}_i\!\!=\!\!
    \begin{cases}
    (\frac{1}{|\mathbf{M}_i^\Phi|}\!\!\sum\limits_{v\in \mathbf{M}_i^\Phi}\!\!\!{f_s(\mathbf{h}_i^\Phi,\mathbf{h}_v^\Phi)},
    \frac{1}{|\mathbf{M}_i^\Phi|}\!\!\sum\limits_{v\in \mathbf{M}_i^\Phi}\!\!\!{f_s(\mathbf{h}_i^\Psi,\mathbf{h}_v^\Psi)})
    ,\emph{if}\ \text{$a_i^{[1]}\!\!=\!\!0$}\\
    (\frac{1}{|\mathbf{M}_i^\Psi|}\!\!\sum\limits_{v\in \mathbf{M}_i^\Psi}\!\!\!{f_s(\mathbf{h}_i^\Psi,\mathbf{h}_v^\Psi)},
    \frac{1}{|\mathbf{M}_i^\Psi|}\!\!\sum\limits_{v\in \mathbf{M}_i^\Psi}\!\!\!{f_s(\mathbf{h}_i^\Phi,\mathbf{h}_v^\Phi)}),
    \emph{if}\ \text{$a_i^{[1]}\!\!=\!\!1$},\\
    \end{cases}
    \end{equation}
where $f_s$ can be an arbitrary similarity  function. Here we use the cosine similarity function $f_s(\mathbf{x},\mathbf{y})=cos(\mathbf{x},\mathbf{y})$. 
$\mathbf{u}_i$ is a two-dimension vector. In order to better present what $\mathbf{u}_i$ stands for, we take $v_1$ and $v_3$ in Fig. \ref{transfer}(b) as an example. 
$v_1$ is an agent-selected node in $\Phi$, i.e., $a_1^{[1]}=0$, and $v_3$ is an agent-selected node in $\Psi$, i.e., $a_3^{[1]}=1$. 
For $\mathbf{u}_1$, 
its first element $u_1^{(1)}$ is the center similarity between $v_1$ and \{$v_4$, $v_5$\} in the distilled network $\Phi$, 
while its second element $u_1^{(2)}$ is the center similarity between $v_1$ and \{$v_4$, $v_5$\} in the guided network $\Psi$.
Similarly, for $\mathbf{u}_3$, $u_3^{(1)}$ measures the center similarity between $v_3$ and \{$v_2$, $v_6$\} in  $\Psi$, and $u_3^{(2)}$ is  the center similarity between $v_3$ and \{$v_2$, $v_6$\} $\Phi$.
In a word, the first element in $\mathbf{u}_i$ measures the center similarity in the distilled network, and the second element measures the center similarity in the guided network.


	Finally, the structure-level state $s_i^{[2]}$ for the the local structure of node $i$ is expressed as:
\begin{equation}\label{s-state}
\setlength{\abovedisplayskip}{0.05in}
\setlength{\belowdisplayskip}{0.05in}
    \mathbf{s}_i^{[2]}=
    CONCAT(\mathbf{s}_i^{[1]},\mathbf{u}_i),
\end{equation}
where $CONCAT$ is the concatenation operation.

    \subsubsection{Structure-level Action} 
    Structure-level action $a_i^{[2]}\in\{0,1\}$ is the second level action that determines which of the structure-level knowledge to be propagated. 
    If $a_i^{[2]}=1$, the agent decides to transfer the knowledge of the local structure encoded in the agent-selected neighborhood set of node $i$, otherwise it will not be transferred.
    Similar to the node-level policy network, the structure-level policy network $\pi_\mathbf{\delta}$ that produces structure-level actions is also comprised of a three-layer MLP with the $tanh$ activation function.

\subsubsection{Structure-level Knowledge Distillation}
We first introduce how to distill structure-level knowledge from $\Phi$ to $\Psi$. The method for distilling from $\Psi$ to $\Phi$ is the same. First, we define the similarity between two agent-selected nodes $i$ and $j$ by:
\begin{align}
 \setlength{\abovedisplayskip}{0.05in}
\setlength{\belowdisplayskip}{0.05in}
    \hat{s}_{ij}^\Phi=\frac{e^{f_s\left(\mathbf{h}_i^\Phi,\mathbf{h}_j^\Phi\right)}}{\sum_{v\in \mathbf{M}_i^\Phi} e^{f_s\left(\mathbf{h}_i^\Phi,\mathbf{h}_v^\Phi\right)}},\ \ 
    \hat{s}_{ij}^\Psi=\frac{e^{f_s\left(\mathbf{h}_i^\Psi,\mathbf{h}_j^\Psi\right)}}{\sum_{v\in \mathbf{M}_i^\Phi} e^{f_s\left(\mathbf{h}_i^\Psi,\mathbf{h}_v^\Psi\right)}},
\end{align}
where $f_s$ is the cosine similarity function.

To transfer structure-level knowledge, we propose a new loss function to be minimized as:
\begin{align}\label{struct1}
\setlength{\abovedisplayskip}{0.05in}
\setlength{\belowdisplayskip}{0.05in}
    \mathcal{L}_{struct}^\Psi=\sum_{i=1}^{N}{(1-a_i^{[1]})a_i^{[2]} KL(\mathbf{\hat{s}}_i^\Phi||\mathbf{\hat{s}}_i^\Psi)},
\end{align}
where $\mathbf{\hat{s}}_i^\Phi=[\hat{s}_{i1}^\Phi, \cdots, \hat{s}_{iC_i^{\Phi}}^\Phi]$ and $\mathbf{\hat{s}}_i^\Psi=[\hat{s}_{i1}^\Psi, \cdots, \hat{s}_{iC_i^{\Phi}}^\Psi]$, $C_i^{\Phi}$ is the size of  $\mathbf{M}_i^\Phi$. $\mathbf{\hat{s}}_i^\Phi$ represents the distribution of the similarities between node $i$ and its agent-selected neighborhoods in $\Phi$, while $\mathbf{\hat{s}}_i^\Psi$ represents the distribution of the similarities between node $i$ and its corresponding neighborhoods in $\Psi$.
If the local structure of node $i$ is decided to transfer, we adopt the $KL$ divergence to make $\mathbf{\hat{s}}_i^\Psi$ match $\mathbf{\hat{s}}_i^\Phi$, so as to transfer structure-level knowledge.
Similarly, we can propose another new loss function for distilling knowledge from $\Psi$ to $\Phi$ as:
\begin{equation}\label{struct2}
\setlength{\abovedisplayskip}{0.05in}
\setlength{\belowdisplayskip}{0.05in}
    \mathcal{L}_{struct}^\Phi=\sum_{i=1}^{N}{a_i^{[1]}a_i^{[2]}KL(\mathbf{\bar{s}}_i^\Psi||\mathbf{\bar{s}}_i^\Phi)},
\end{equation}
where $\mathbf{\bar{s}}_i^\Phi=[\bar{s}_{i1}^\Phi, \cdots, \bar{s}_{iC_i^{\Psi}}^\Phi]$ and $\mathbf{\bar{s}}_i^\Psi=[\bar{s}_{i1}^\Psi, \cdots, \bar{s}_{iC_i^{\Psi}}^\Psi]$, $C_i^{\Psi}$ is the size of  $\mathbf{M}_i^\Psi$. $\bar{s}_{ij}^\Phi$ and $\bar{s}_{ij}^\Psi$ are defined as:
\begin{align}
\setlength{\abovedisplayskip}{0.05in}
\setlength{\belowdisplayskip}{0.05in}
    \bar{s}_{ij}^\Phi=\frac{e^{f_s\left(\mathbf{h}_i^\Phi,\mathbf{h}_j^\Phi\right)}}{\sum_{v\in \mathbf{M}_i^\Psi} e^{f_s\left(\mathbf{h}_i^\Phi,\mathbf{h}_v^\Phi\right)}}, \ \ 
    \bar{s}_{ij}^\Psi=\frac{e^{f_s\left(\mathbf{h}_i^\Psi,\mathbf{h}_j^\Psi\right)}}{\sum_{v\in \mathbf{M}_i^\Psi} e^{f_s\left(\mathbf{h}_i^\Psi,\mathbf{h}_v^\Psi\right)}}.
\end{align}

By jointly minimizing (\ref{struct1}) and (\ref{struct2}), we can dynamically exchange structure-level knowledge  between $\Phi$ and $\Psi$.

\subsection{Optimizations}
In this section, we introduce the optimization procedure of our method.

\subsubsection{Reward} 
Following \cite{wang2019minimax}, our actions are sampled in batch, and obtain the delayed reward after two GNNs being updated according to a batch of sequential actions.
Similar to \cite{ liang2021reinforced}, we utilize the performance of the models after being updated as the reward.
We use the negative value of the cross entropy loss to measure the performance of the models as in \cite{yuan2021reinforced, liang2021reinforced}, defined as:
 \begin{equation}\label{reward}
 \setlength{\abovedisplayskip}{0.05in}
\setlength{\belowdisplayskip}{0.05in}
     R_i\!=\!-\frac{\sum\limits_{u\in \mathbf{B}}{(\mathcal{L}_{CE}^\Phi(u){+\mathcal{L}}_{CE}^\Psi(u))}}{\left|\mathbf{B}\right|}-\gamma\frac{\sum\limits_{v\in {\mathcal{N}_i}}{(\mathcal{L}_{CE}^\Phi(v){+\mathcal{L}}_{CE}^\Psi(v))}}{\left|{\mathcal{N}_i}\right|},
 \end{equation}
where $\gamma$ is a hyper-parameter.
$R_i$ is the reward for the action taken at node $i$, and $\mathbf{B}$ is a batch set of nodes from the training set.
The reward for an action  $a_i$ consists of two parts: The first part is the average performance for a batch of nodes,  measuring the global effects that the action $a_i$ brings on the GNN model; The second part is the average performance of the neighborhoods of node $i$,  in order to model the local effects of $a_i$. 

\subsubsection{Optimization for Policy Networks} 
\label{policy_opt}
Following previous studies about hierarchical reinforcement learning \cite{liu2021rmm}, the gradient of expected cumulative reward $\mathrm{\nabla}_{\theta,\delta}J$ could be computed as follows:
\begin{equation}\label{grad}
\setlength{\abovedisplayskip}{0.05in}
\setlength{\belowdisplayskip}{0.05in}
\mathrm{\nabla}_{\theta,\delta}J\!=\!
\frac{1}{|\mathbf{B}|}\!
\sum_{i\in\mathbf{B}\!}{\!(R_i\!-\!b_i)\mathrm{\nabla}_{\theta,\delta}\!\log(\pi_\theta(\mathbf{s}_i^{[1]}\!,a_i^{[1]})\pi_\delta(\mathbf{s}_i^{[2]}\!,a_i^{[2]})}),
\end{equation}
where $\theta$, $\delta$ is the learned parameters of the node-level policy network and structure-level policy network, respectively. Similar to \cite{lai2020policy}, to speed up convergence and reduce variance ,  we also add a baseline reward $b_i$ that is the rewards at node $i$ in the last epoch. The motivation behind this is to encourage the agent to achieve better performance than that of the last epoch. Finally, we update the parameters of policy networks by gradient ascent \cite{williams1992simple} as:
\begin{equation}\label{update}
    \theta\gets\theta+\eta\mathrm{\nabla}_\theta J,\ \   \delta\gets\delta+\eta\mathrm{\nabla}_\delta J,
\end{equation}
where $\eta$ is the learning rate for reinforcement learning.

\begin{algorithm}[h]
\label{alg}
  \caption{The training procedure of FreeKD.}  
  \begin{algorithmic}[1]  
    \Require 
    graph $\mathbf{G=(V,E,X)}$, label set $\mathbf{\mathcal{Y}}$, epoch number $L$;
    \Ensure
    the predicted classes of nodes in the GNN models $\Phi$ and $\Psi$, the trained parameters of  $\Phi$ and  $\Psi$;
    \State Initialize  $\Phi$ and  $\Psi$;
    \State Initialize the policy networks in reinforcement learning;
    \For{each epoch $k$ in $1$ to $L$}  
        \For{each batch in epoch $k$}  
		    \State calculate the cross entropy losses $\mathcal{L}_{CE}^\Phi$, $\mathcal{L}_{CE}^\Psi$;
		    \State calculate node-level states for a batch of nodes;
		    \State  sample node-level actions by $a_i^{[1]}\!\!\sim\!\!\pi_\mathbf{\theta}(\!\mathbf{s}_i^{[1]}\!,\!a_i^{[1]}\!)$;
		    \State derive structure-level states;
		    \State  sample structure-level actions by $a_i^{[2]}\!\!\sim\!\!\pi_\mathbf{\delta}(\!\mathbf{s}_i^{[2]}\!,\!a_i^{[2]}\!)$;
		    \State store states and actions to history buffer $H$;
		    \State  calculate $\mathcal{L}_{node}^\Phi$ and $\mathcal{L}_{node}^\Psi$;
		    \State calculate $\mathcal{L}_{struct}^\Phi$ and $\mathcal{L}_{struct}^\Psi$;
		    \State  calculate the overall losses $\mathcal{L}^\Phi$, $\mathcal{L}^\Psi$;
		    \State update the parameters of  $\Phi$ by minimizing $L^\Phi$;
		    \State update the parameters of $\Psi$ by minimizing $\mathcal{L}^\Psi$;
		    \For{each state and action in buffer $H$}
            	\State calculate the delayed rewards;
        		\State update parameters of the policy networks;
            \EndFor
        \EndFor  
    \EndFor     
  \end{algorithmic}  
\end{algorithm}  

\subsubsection{Optimization for GNNs}
We minimize the following loss functions for optimizing  $\Phi$ and $\Psi$, respectively:
\begin{equation}\label{loss1}
\setlength{\abovedisplayskip}{0.05in}
\setlength{\belowdisplayskip}{0.05in}
    \mathcal{L}^\Phi=\mathcal{L}_{CE}^\Phi+\mu \mathcal{L}_{node}^\Phi+\rho L_{struct}^\Phi
\end{equation}
\begin{equation}\label{loss2}
    \mathcal{L}^\Psi=\mathcal{L}_{CE}^\Psi+\mu \mathcal{L}_{node}^\Psi+\rho L_{struct}^\Psi,   
\end{equation}
\begin{equation}\label{loss3}
    \mathcal{L}_{sum} =  \mathcal{L}^\Phi + \mathcal{L}^\Psi
\end{equation}
where $\mathcal{L}_{CE}^\Phi$, $\mathcal{L}_{CE}^\Psi$ are the cross entropy losses for  $\Phi$ and  $\Psi$, respectively. $\mathcal{L}_{node}^\Phi$ and $\mathcal{L}_{node}^\Psi$ are two node-level knowledge distillation losses. $\mathcal{L}_{struct}^\Phi$ and $\mathcal{L}_{struct}^\Psi$ are two structure-level distillation losses. 
$\mathcal{L}_{sum}$ is the overall loss for the two GNNs.
$\mu$ and $\rho$ are two trade-off parameters.

The pseudo-code of the our FreeKD training procedure is listed in Algorithm 1. The GNNs and the agent closely interact with each other when training. For each batch, we first calculate the cross entropy loss for training GNNs. The node-level states for a batch of nodes are then calculated and feed into the node-level policy network. After that, we sample node-level actions from the policy probabilities produced by the agent to decide the directions of knowledge distillation between two GNNs. Then, the agent receives structure-level states from environment and produces structure-level actions to decide which of the local structures to be propagated.
The two-level states and actions are stored in the history buffer. Next, we train the two GNNs with the overall loss. After that, for the stored states and actions, we calculate the delayed rewards according to the performance of GNNs and update the policy network with gradient ascent. The GNNs and the agent are learned together and mutually improved.

\begin{figure}
  \centering
  \includegraphics[width=1.0\linewidth]{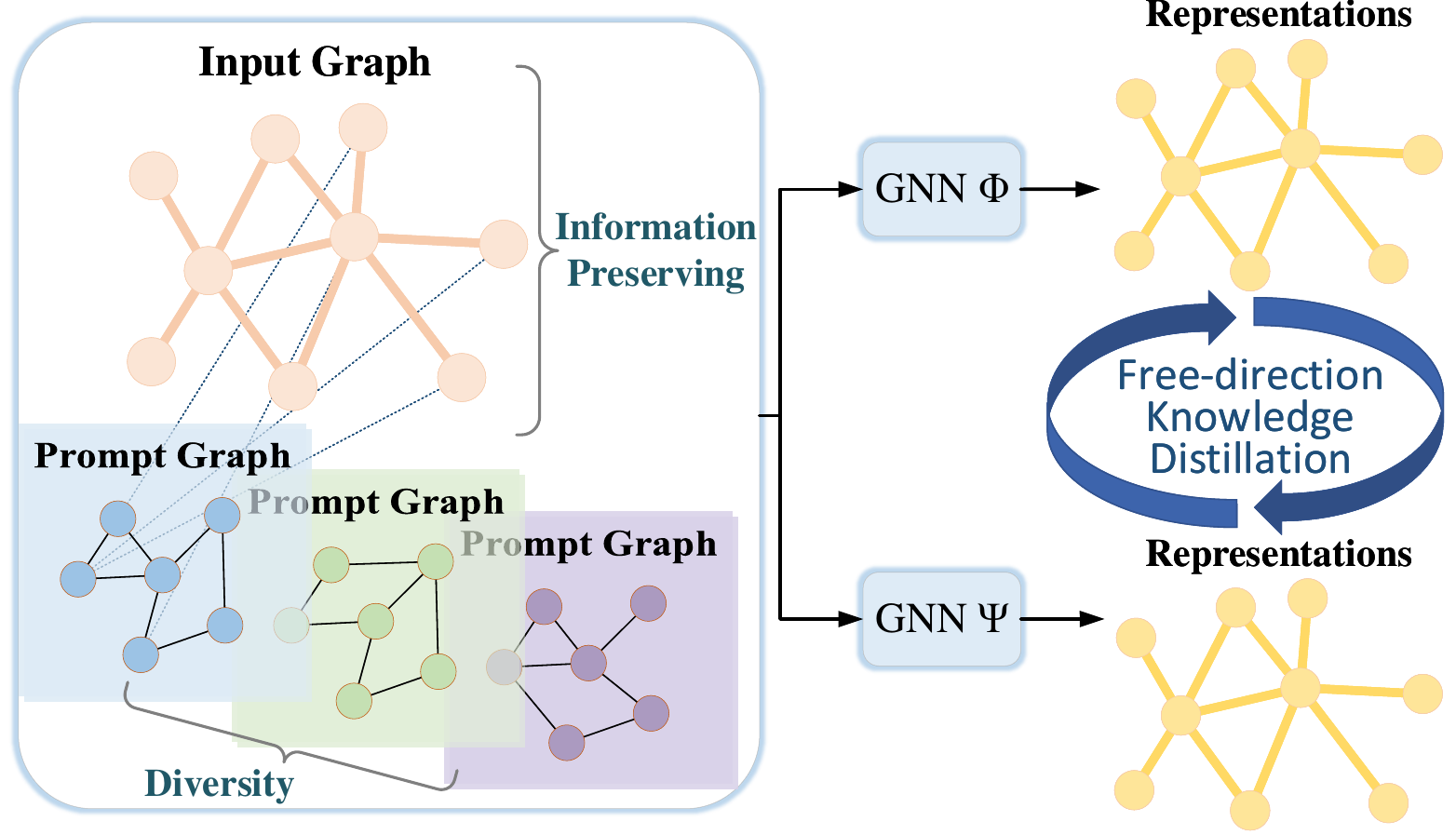}
  \caption{
    An illustration of our proposed FreeKD-Prompt. Various learned prompt graphs are alternately inserted to the input graph for exchanging varied knowledge between two GNNs.  
  }
  \label{freekd-p}
\end{figure}

\subsection{Prompt-enhanced Knowledge Distillation}
In this section, we present FreeKD-Prompt, a new approach aimed at promoting diverse knowledge exchange between two GNNs through prompt-based graph augmentations, motivated by \cite{sun2023all}. 
The architecture of FreeKD-Prompt is illustrated in Figure \ref{freekd-p}.
We represent our prompt in the form of a graph and seamlessly integrate it into the input graph to facilitate augmentations.
Then, we formulate two loss functions to encourage the learned augmented graphs to be both undistorted and diverse, thus allowing for effective exchange of varied knowledge from different augmented perspectives. 
In the next, we will begin by introducing the design of the prompt-based graph augmentation, followed by the optimization of the prompt graph.

\subsubsection{Graph Augmentation via Prompt Graph}
We first introduce the notations of the prompt graph. Similar to \cite{sun2023all}, we denote the prompt graph as $\mathbf{G}^p = (\mathbf{V}^p, \mathbf{X}^p, \mathbf{E}^p)$, where $\mathbf{V}^p=\{v^p_i\}^{P}_{i=1}$ is the set of prompt token nodes, $\mathbf{X}^p=\{\mathbf{x}^p_i\}^{P}_{i=1}$ is the set of learnable token features, $\mathbf{E}^p$ is the edge set between prompt tokens. 
$P$ is the number of prompt token nodes and we usually set it as a small number (e.g. $P=100$) in practice for parameter efficiency, following \cite{sun2023all}.
Each prompt token node $v^p_i$ is represented by a parameterized token vector $\mathbf{x}^p_i \in \mathbb{R}^{1\times d}$, where $d$ is the dimension matching the node feature in the input graph. 
We formulate the edges between prompt tokens as follows:
\begin{equation}
    \mathbf{E}^p=\{(v^p_i,v^p_j)\ |\ f_s(\mathbf{x}^p_i,\mathbf{x}^p_j) \ge \varphi_p\ \mathrm{and}\ v^p_i,v^p_j \in \mathbf{V}^p\},
\end{equation}
where $f_s$ is the cosine similarity function and $\varphi_p$ is a threshold of the edge formation between prompt nodes. 

Subsequently, to insert the prompt graph to the input graph, we introduce a similar function for forming edges between the prompt graph and the input graph:
\begin{equation}
    \mathbf{E}^c=\{(v^p_i,v_j)\ |\ f_s(\mathbf{x}^p_i,\mathbf{x}_j) \ge \varphi_c\ \mathrm{and}\ v^p_i \in \mathbf{V}^p, v_j \in \mathbf{V} \},
\end{equation}
where $v_j \in \mathbf{V}$ is the node in the input graph with feature $\mathbf{x}_j \in \mathbf{X}$ and $\varphi_c$ is a threshold of the edge formation between prompt nodes and input graph nodes. $\mathbf{E}^c$ represents the edge set containing edges between the prompt nodes and the input graph nodes. 

In this paper, we adopt a dynamic threshold value $\varphi_c$ instead of a fixed one. This is because a fixed threshold value may cause the instability of the number of edges in $\mathbf{E}^c$ during training, which might make the training process unstable. For instance, it could result in zero edges connecting the prompt graph and the input graph. Thus, we set $\varphi_c$ to be the similarity value ranked at the top $T_{\varphi_c}$ percent among all similarity values between the prompt nodes and the input graph nodes. Similarly, $\varphi_p$ is also set as a dynamic threshold using the same approach.

Finally, after the edges connecting the prompt graph $\mathbf{G}^p$ and the input graph $\mathbf{G}$ are established, we derive an augmented graph denoted as $\mathbf{G}^m_i$. This augmentation process can be represented as $\mathbf{G}^m_i = \Omega(\mathbf{G},\mathbf{G}^p)$. 

To facilitate the distillation of different knowledge, we set up a group of prompt graphs $\{\mathbf{G}^p_i\}^M_{i=1}$ to produce a group of augmented graphs by $\mathbf{G}^m_i = \Omega(\mathbf{G},\mathbf{G}^p_i)$, where $M$ is the number of prompt graphs. By alternately inputting different $\mathbf{G}^m_i$ to the GNN model, we can derive different representations of nodes in the input graph via message passing for distilling varied knowledge. In the next, we will introduce how to learn the prompt graphs.

\subsubsection{Optimizations of Prompt Graphs}

As aforementioned, the derived augmented graph should retain the semantic information of the original graph to prevent drastic semantic changes. 
Motivated by \cite{velivckovic2018deep}, we devise an information persevering loss based on mutual information maximization to encourage the augmented graph derived by the prompts to maintain the original information. 
To achieve this, we expect that each node representation in the augmented graph encompasses the semantic information in the original graph from local level to global level.
Specifically, we intend to maximize the mutual information between the node representation in the augmented graph and the global graph representations as well as the local neighborhood representations in the original graph.

Therefore, our objective function can be formulated as:
\begin{equation}
    J(\mathbf{G},\mathbf{G}^m_i)\!=\!max\! \sum_{k=1}^{N} [I(h_k(\mathbf{G}^m_i),g^s\!(\mathbf{G})) + I(h_k(\mathbf{G}^m_i),g_k^s\!(\mathbf{G}))],
    \label{mimax}
\end{equation}
where $N$ is the number of nodes in $\mathbf{G}$ and $I$ denotes the mutual information. $h_k(\mathbf{G}^m_i)$ represents the output representations of node $k$ obtained by the GNN model taken $\mathbf{G}^m_i$ as input. 
For convenience, here we directly use one of the in-training GNN models (e.g. GNN $\Phi$) for this encoding. 
$g^s\!(\mathbf{G})$ denotes the global graph representation, calculated by averaging all the node representations obtained by the GNN model with $\mathbf{G}$ as input. On the other hand, $g^s_k\!(\mathbf{G})$ refers to the local neighborhood representation, calculated by averaging the node representations within the $1$-hop neighborhood of node $k$. 

By maximizing Eq. (\ref{mimax}), each node representation in $\mathbf{G}^m_i$ is encouraged to preserve information in the original input graph from local to global receptive fields, thus avoiding the drastic semantic changes. 
Since it's intractable to directly optimize the mutual information, we adopt an approximation approach in  \cite{oord2018representation,velivckovic2018deep} to maximize the mutual information in $J(\mathbf{G},\mathbf{G}^m_i)$ as follows: 
 \begin{align} \nonumber
\mathcal{L}_{info}(\mathbf{G},\mathbf{G}^m_i) =& -\sum_{k=1}^{N} (\log \mathcal{D}(h_k(\mathbf{G}^m_i),g^s\!(\mathbf{G}))) \\ \nonumber
     & -\sum_{k=1}^{N} (\log (1-\mathcal{D}(h_k(\widetilde{\mathbf{G}}^m_i),g^s\!(\mathbf{G})))) \\ \nonumber
     & -\sum_{k=1}^{N} (\log \mathcal{D}(h_k(\mathbf{G}^m_i),g_k^s\!(\mathbf{G}))) \\
    & -\sum_{k=1}^{N} (\log (1-\mathcal{D}(h_k(\widetilde{\mathbf{G}}^m_i),g_k^s\!(\mathbf{G})))),
 \end{align}
 where $\mathcal{L}_{info}(\mathbf{G},\mathbf{G}^m_i)$ is the loss for maximizing $J(\mathbf{G},\mathbf{G}^m_i)$ approximately.
 $\mathcal{D}$ represents a discriminator that is expected to output a higher probability score for the positive pair $(h_k(\mathbf{G}^m_i),g^s\!(\mathbf{G}))$ and a lower probability score for the negative pair $(h_k(\widetilde{\mathbf{G}}^m_i),g^s\!(\mathbf{G}))$. 
 Following \cite{velivckovic2018deep}, the negative pair is derived by replacing the node representation $h_k(\mathbf{G}^m_i)$ in the positive pair by the node representation $h_k(\widetilde{\mathbf{G}}^m_i)$ from the corrupted graph. 
 The corrupted graph $\widetilde{\mathbf{G}}^m_i$ is generated by conducting row-wise shuffling on the node feature matrix. The discriminator $\mathcal{D}$ is implemented as a bilinear score function:
 \begin{equation}
     \mathcal{D}(h_k(\mathbf{G}^m_i),g^s\!(\mathbf{G})) = \sigma ({h_k(\mathbf{G}^m_i)}^T W_{\mathcal{D}} g^s\!(\mathbf{G})),
 \end{equation}
where $W_{\mathcal{D}}$ is the weight matrix and $\sigma(\cdot)$ is the sigmoid activation function.

In addition, to ensure the diversity of augmented graphs, we devise a diversity loss $\mathcal{L}_{div}$ as follows:
\begin{align} \nonumber
    \mathcal{L}_{div} = & -\sum_{i=1}^{M} \sum_{j=1}^{M}  \sum_{k=1}^{P} f_s(\mathbf{x}^p_{i,k},\mathbf{x}^p_{j,k})) \\
    & -\sum_{i=1}^{M} \sum_{j=1}^{M}  \sum_{k=1}^{N} f_s(h_k(\mathbf{G}^m_i),h_k(\mathbf{G}^m_j)) ,
\end{align}
where  $f_s$ is the cosine similarity function and $\mathbf{x}^p_{i,k}$ denotes the $k^{th}$ prompt token in the $i^{th}$ prompt graph $\mathbf{G}^p_i$.
This loss assesses diversity from two perspectives. The first term directly quantifies the disparity among the prompt tokens across various prompt graphs. The second term measures the distinction in semantic node representations across different augmented graphs.

By combining the above two losses, the overall loss for learning the graph prompt can be written as follows:
\begin{equation}
    \mathcal{L}_p = \sum_{i=1}^{M} \mathcal{L}_{info}(\mathbf{G},\mathbf{G}^m_i)+\beta \mathcal{L}_{div},
    \label{ploss}
\end{equation}
where $\beta$ serves as a trade-off hyper-parameter to balance information preservation and diversity. By optimizing this loss, we can obtain augmented graphs that are both diverse and undistorted.

Finally, we input the derived augmented graph in turn to facilitate the exchange of varied knowledge  between two GNNs. The training loss for the two GNNs are as follows:
\begin{equation}
    \mathcal{L}_{pkd} = \sum_{i=0}^M \mathcal{L}_{sum}(\mathcal{G}^\Phi(\mathbf{G}^m_i),\mathcal{G}^\Psi(\mathbf{G}^m_i)),
\end{equation}
where $\mathcal{L}_{sum}(\mathcal{G}^\Phi(\mathbf{G}^m_i),\mathcal{G}^\Psi(\mathbf{G}^m_i))$ denotes the overall loss of two GNNs $\Phi$ and $\Psi$ by using FreeKD  when they both take as input the augmentation view $\mathbf{G}^m_i$. Note that $\mathbf{G}^m_0$ denotes the original input graph without augmentations. 
In practice, we alternately optimize $\mathcal{L}_p$, $\mathcal{L}_{pkd}$, and the reward $\mathrm{\nabla}_{\theta,\delta}J$ for training the prompts,  the GNNs and the agent respectively. In this way, our FreeKD-Prompt enables the exchange of diverse knowledge from different augmentation views between two GNNs, leading to a more comprehensive knowledge transfer.

\subsection{Extension to Multiple GNNs}

In the previous sections, we primarily focus on  free-direction knowledge distillation between two GNN models. To further enhance our FreeKD, we introduce FreeKD++ to facilitate free-direction knowledge transfer among multiple GNNs The motivations behind FreeKD++ are as follows: 
each GNN model may have its own strengths and weaknesses in capturing different aspects of the graph data. By involving multiple GNNs as potential knowledge sources in the knowledge transfer process, we can harness the collective intelligence of these models.  If one GNN fails to capture certain aspects or encounters limitations, other GNNs can compensate for it. This can enhance the effectiveness of the knowledge distillation process, reducing the risk of over-reliance on a single model and mitigating the impact of individual model weaknesses.
In the next, we will introduce our FreeKD++.

For simplification, here we denote $\mathcal{G}^i$ as $i^{th}$ GNN model and $K$ as the number of GNNs.
We exchange knowledge between each pair of GNNs within the set of $K$ GNNs. This process is defined as follows:
\begin{equation}
    \mathcal{L}_{m} = \sum_{i=1}^K \sum_{j=1}^K  \mathcal{L}_{sum}(\mathcal{G}^i(\mathbf{G}),\mathcal{G}^j(\mathbf{G})),
\end{equation}
where $\mathcal{L}_{sum}(\mathcal{G}^i(\mathbf{G}),\mathcal{G}^j(\mathbf{G}))$ denotes the overall loss of two GNNs $\mathcal{G}^i$ and $\mathcal{G}^j$ by using FreeKD  when they both take as input $\mathbf{G}$. By optimizing $\mathcal{L}_m$, we enable the exchange of knowledge among multiple GNNs. It is worth noting that all of these GNNs share a same agent, and the optimization of the agent follows the same procedure as described in Section \ref{policy_opt}.

\begin{figure}
  \centering
  \includegraphics[width=0.9\linewidth]{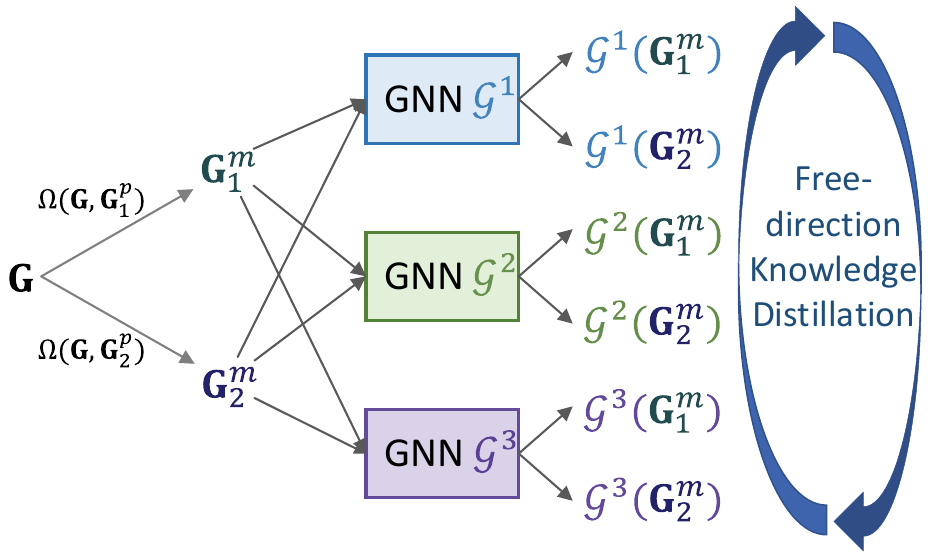}
    \vspace{-0.2in}
  \caption{
    An illustration of FreeKD-Prompt++. Multiple GNNs exchange varied knowledge under different prompt-based augmentation inputs.
  }
    \vspace{-0.2in}
  \label{freekd++}
\end{figure}

In addition, we further introduce FreeKD-Prompt++ by incorporating the prompt-based distillation into FreeKD++, to facilitate the exchange of varied knowledge between multiple GNNs. The framework of FreeKD-Prompt++ is illustrated in Fig. \ref{freekd++}. Similarly, the loss $\mathcal{L}_m$ for $K$ GNNs can be written as:
\begin{equation}
    \mathcal{L}_{m} = \sum_{i=1}^K \sum_{j=1}^K \sum_{k=0}^M  \mathcal{L}_{sum}(\mathcal{G}^i(\mathbf{G}^m_k),\mathcal{G}^j(\mathbf{G}^m_k)),
\end{equation}
where $\mathbf{G}^m_k$ is the $k^{th}$ prompt-based augmentation view and $\mathbf{G}^p_0$ denotes the original graph without augmentations.

\section{EXPERIMENTS}
To verify the effectiveness of our proposed methods, we perform the experiments on five benchmark datasets of different domains and on GNNs of different architectures. 

\subsection{Experimental Setups}
\subsubsection{Datasets} 
 We use five widely used benchmark datasets to evaluate our methods.
 Cora \cite{sen2008collective} and Citeseer \cite{sen2008collective} are two citation datasets where  nodes represent documents and   edges represent citation relationships. 
Chameleon \cite{rozemberczki2021multi} and Texas \cite{pei2020geom} are two web network datasets where  nodes stand for web pages and  edges show their hyperlink relationships. 
The PPI dataset \cite{hamilton2017inductive} consists of 24 protein–protein interaction graphs, corresponding to different human tissues.
Table \ref{dataset} summarizes the statistics of the five datasets.
Following \cite{chen2018fastgcn} and \cite{huang2018adaptive}, we use 1000 nodes for testing, 500 nodes for validation, and the rest for training on the Cora and Citeseer datasets.
For Chameleon and Texas datasets, we randomly split nodes of each class into 60\%, 20\%, and 20\% for training, validation and testing respectively, following \cite{pei2020geom} and \cite{chen2020simple}.
For the PPI dataset, we use 20 graphs for training, 2 graphs for validation, and 2 graphs for testing, as in  \cite{chen2020simple}.
Following previous works \cite{velivckovic2017graph,pei2020geom} , we study the transductive setting on the first four datasets, and the inductive setting on the PPI dataset.
In the tasks of transductive setting, we predict the labels of the nodes observed during training, whereas in the task of inductive setting, we predict the labels of nodes in never seen graphs before. Following \cite{chen2018fastgcn,hamilton2017inductive}, we use the Micro-F1 score as the evaluation measure throughout the experiment.

\subsubsection{Baselines} In the experiment, we adopt three popular GNN models, GCN \cite{kipf2016semi}, GAT \cite{velivckovic2017graph}, GraphSAGE \cite{hamilton2017inductive}, as our basic models in our method.
Our framework aims to promote the performance of these GNN models. Thus, these three GNN models can be used as our baselines. We briefly introduce these three GNNs as follows:

\begin{itemize}
\item GCN \cite{kipf2016semi} is a convolutional neural network designed for graph-structured data. It performs convolution operations to capture structural patterns in graphs.
\item GraphSAGE \cite{hamilton2017inductive} enables efficiently inductive learning on large-scale graphs by proposing an efficient sample strategy to aggregate neighbor nodes.
\item GAT \cite{velivckovic2017graph} utilizes the self-attention mechanism to assign varying weights to neighbors, so as to effectively aggregate neighbor information.
\end{itemize}

Since we propose a free-direction knowledge distillation framework, we also compare  with five typical knowledge distillation approaches proposed recently, including KD \cite{hinton2015distilling}, LSP \cite{yang2020distilling}, CPF \cite{yang2021extract},  GNN-SD \cite{chen2020self}, RDD \cite{zhang2020reliable} and G-CRD \cite{joshi2022representation},  to further verify the effectiveness of our method. We briefly introduce these knowledge distillation methods as follows:

\begin{itemize}
\item KD \cite{hinton2015distilling} is a vanilla knowledge distillation method that directly transfers knowledge from soft labels.
\item LSP \cite{yang2020distilling} distills the topological structure knowledge from a pre-trained deeper teacher GNN to a shallower student GNN.
\item CPF \cite{yang2021extract} combines the parameterized label propagation and the feature transformation module in the student to improve the distillation performance.
\item GNN-SD \cite{chen2020self} attempts to distill knowledge from the shallow layers to the deep layers in the GNN model, so as to alleviate the over-smoothing issue.
\item  RDD \cite{zhang2020reliable} is a semi-supervised knowledge distillation method which learns a complicated teacher GNN model by ensemble learning for distillation.
\item G-CRD \cite{joshi2022representation} proposes a distillation method based on contrastive learning to encourage the student GNN preserve the global topology knowledge learned by the teacher GNN.
\end{itemize}

\begin{table}
\centering
\vspace{-0.1in}
\caption{Dataset statistics.}
\vspace{-0.1in}
\renewcommand\arraystretch{1.1}
 \setlength{\tabcolsep}{1.3mm}{
\begin{tabular}{@{}cccccc@{}}
\toprule
Dataset   & \# Graphs & \# Nodes & \# Edges & \# Features & \# Classes \\ \midrule
Cora      & 1         & 2708     & 5429     & 1433        & 7          \\
Citeseer  & 1         & 3327     & 4732     & 3703        & 6          \\
Chameleon & 1         & 2277     & 36101    & 2325        & 4          \\
Texas     & 1         & 183      & 309      & 1703        & 5          \\
PPI       & 24        & 56944    & 818716   & 50          & 121        \\ \bottomrule
\end{tabular}
}
\vspace{-0.2in}
 \label{dataset}
\end{table}

\begin{table*}
    \centering
    \caption{Results (\%) of the compared approaches for node classification in the transductive settings on the Cora, Chameleon, Citeseer, and Texas datasets. The values in the brackets denote the performance improvement of our FreeKD and FreeKD-Prompt over the corresponding baselines. Here, we denote GraphSAGE as SAGE for short.}
    \vspace{-0.1in}
    \renewcommand\arraystretch{1.1}
    \setlength{\tabcolsep}{1.3mm}{%
    \begin{tabular}{@{}ccccccccccc@{}}
    \toprule
                                & \multicolumn{2}{c}{}             & \multicolumn{2}{c}{Cora}                         & \multicolumn{2}{c}{Chameleon}                     & \multicolumn{2}{c}{Citeseer}                    & \multicolumn{2}{c}{Texas}             \\ \midrule
    \multicolumn{1}{c|}{Method} & \multicolumn{2}{c|}{Basic Model}     & \multicolumn{2}{c|}{F1 Score ($\uparrow$Impv.)}           & \multicolumn{2}{c|}{F1 Score ($\uparrow$Impv.)}            & \multicolumn{2}{c|}{F1 Score ($\uparrow$Impv.)}           & \multicolumn{2}{c}{F1 Score ($\uparrow$Impv.)} \\
    \multicolumn{1}{c|}{}       &  $\Phi$ & \multicolumn{1}{c|}{ $\Psi$} & $\Phi$         & \multicolumn{1}{c|}{$\Psi$}         & $\Phi$         & \multicolumn{1}{c|}{$\Psi$}         & $\Phi$         & \multicolumn{1}{c|}{$\Psi$}         & $\Phi$              & $\Psi$              \\
    \midrule
    \multicolumn{1}{c|}{GCN} & -  & \multicolumn{1}{c|}{-}     & 85.12        & \multicolumn{1}{c|}{-}             & 33.09        & \multicolumn{1}{c|}{-}             & 75.42        & \multicolumn{1}{c|}{-}             & 57.57             &       -            \\
    \multicolumn{1}{c|}{SAGE} & - & \multicolumn{1}{c|}{-}     & 85.36        & \multicolumn{1}{c|}{-}             & 48.77        & \multicolumn{1}{c|}{-}             & 76.56        & \multicolumn{1}{c|}{-}             & 76.22             &          -         \\
    \multicolumn{1}{c|}{GAT} & -  & \multicolumn{1}{c|}{-}     & 85.45        & \multicolumn{1}{c|}{-}             & 40.29       & \multicolumn{1}{c|}{-}             & 75.66         & \multicolumn{1}{c|}{-}             & 57.84             &        -           \\
    \midrule
    \multicolumn{1}{c|}{FreeKD}  & GCN  & \multicolumn{1}{c|}{GCN}  & 86.53(\textbf{$\uparrow$1.41}) & \multicolumn{1}{c|}{86.62(\textbf{$\uparrow$1.50})} & 37.48(\textbf{$\uparrow$4.39})  & \multicolumn{1}{c|}{37.79(\textbf{$\uparrow$4.70})} & 77.28(\textbf{$\uparrow$1.86}) & \multicolumn{1}{c|}{77.33(\textbf{$\uparrow$1.91})} & 60.00(\textbf{$\uparrow$2.43})      & 60.81(\textbf{$\uparrow$3.24})      \\
    \multicolumn{1}{c|}{FreeKD}  & SAGE & \multicolumn{1}{c|}{SAGE} & 86.41(\textbf{$\uparrow$1.05}) & \multicolumn{1}{c|}{86.55(\textbf{$\uparrow$1.19})} & 49.89(\textbf{$\uparrow$1.12}) & \multicolumn{1}{c|}{49.78(\textbf{$\uparrow$1.01})} &  77.78(\textbf{$\uparrow$1.22}) & \multicolumn{1}{c|}{ 77.58(\textbf{$\uparrow$1.02})} & 77.84(\textbf{$\uparrow$1.62})      & 77.57(\textbf{$\uparrow$1.35})      \\
    \multicolumn{1}{c|}{FreeKD}  & GAT  & \multicolumn{1}{c|}{GAT}  & 86.46(\textbf{$\uparrow$1.01}) & \multicolumn{1}{c|}{86.68(\textbf{$\uparrow$1.23})} & 44.32(\textbf{$\uparrow$4.03}) & \multicolumn{1}{c|}{44.10(\textbf{$\uparrow$3.81})} &  77.13(\textbf{$\uparrow$1.47}) & \multicolumn{1}{c|}{77.42(\textbf{$\uparrow$1.76})} & 61.35(\textbf{$\uparrow$3.51})      & 61.27(\textbf{$\uparrow$3.43})      \\
    \multicolumn{1}{c|}{FreeKD}  & GCN  & \multicolumn{1}{c|}{GAT}  & 86.65(\textbf{$\uparrow$1.53}) & \multicolumn{1}{c|}{86.72(\textbf{$\uparrow$1.27})} & 35.61(\textbf{$\uparrow$2.52}) & \multicolumn{1}{c|}{43.44(\textbf{$\uparrow$3.15})} &  77.39(\textbf{$\uparrow$1.97}) & \multicolumn{1}{c|}{77.58(\textbf{$\uparrow$1.92})} & 60.81(\textbf{$\uparrow$3.24})      & 61.35(\textbf{$\uparrow$3.51})      \\
    \multicolumn{1}{c|}{FreeKD}  & GCN  & \multicolumn{1}{c|}{SAGE} & 86.26(\textbf{$\uparrow$1.14}) & \multicolumn{1}{c|}{86.76(\textbf{$\uparrow$1.40})} & 36.73(\textbf{$\uparrow$3.64}) & \multicolumn{1}{c|}{49.93(\textbf{$\uparrow$1.16}) } &  77.08(\textbf{$\uparrow$1.66}) & \multicolumn{1}{c|}{77.68(\textbf{$\uparrow$1.12})} & 60.54(\textbf{$\uparrow$2.97})      & 78.11(\textbf{$\uparrow$1.89})      \\
    \multicolumn{1}{c|}{FreeKD}  & GAT  & \multicolumn{1}{c|}{SAGE} & 86.67(\textbf{$\uparrow$1.22}) & \multicolumn{1}{c|}{86.84(\textbf{$\uparrow$1.48})} & 43.82(\textbf{$\uparrow$3.53})  & \multicolumn{1}{c|}{ 49.85(\textbf{$\uparrow$1.08})} & 77.24(\textbf{$\uparrow$1.58}) & \multicolumn{1}{c|}{77.62(\textbf{$\uparrow$1.06}) } & 62.16(\textbf{$\uparrow$4.32})      & 77.30(\textbf{$\uparrow$1.08})      \\ 
    \midrule
    \multicolumn{1}{c|}{FreeKD-Prompt}  & GCN  & \multicolumn{1}{c|}{GCN}  & 87.33(\textbf{$\uparrow$2.21}) & \multicolumn{1}{c|}{87.46(\textbf{$\uparrow$2.34})} & 40.04(\textbf{$\uparrow$6.95})  & \multicolumn{1}{c|}{40.07(\textbf{$\uparrow$6.98})} & 78.53(\textbf{$\uparrow$3.11}) & \multicolumn{1}{c|}{78.54(\textbf{$\uparrow$3.12})} & 65.41(\textbf{$\uparrow$7.84})      & 66.76(\textbf{$\uparrow$9.19})     \\
    \multicolumn{1}{c|}{FreeKD-Prompt}  & SAGE & \multicolumn{1}{c|}{SAGE}  & 87.33(\textbf{$\uparrow$1.97}) & \multicolumn{1}{c|}{87.45(\textbf{$\uparrow$2.09})} & 50.07(\textbf{$\uparrow$1.30})  & \multicolumn{1}{c|}{50.00(\textbf{$\uparrow$1.23})} & 78.39(\textbf{$\uparrow$1.83}) & \multicolumn{1}{c|}{78.39(\textbf{$\uparrow$1.83})} & 81.08(\textbf{$\uparrow$4.86})      & 79.19(\textbf{$\uparrow$2.97})      \\
    \multicolumn{1}{c|}{FreeKD-Prompt}  & GAT  & \multicolumn{1}{c|}{GAT}  & 87.44(\textbf{$\uparrow$1.99}) & \multicolumn{1}{c|}{87.27(\textbf{$\uparrow$1.82})} & 45.68(\textbf{$\uparrow$5.39})  & \multicolumn{1}{c|}{45.35(\textbf{$\uparrow$5.06})} & 78.05(\textbf{$\uparrow$2.39}) & \multicolumn{1}{c|}{78.00(\textbf{$\uparrow$2.34})} & 64.60(\textbf{$\uparrow$6.76})      & 64.87(\textbf{$\uparrow$7.03})      \\
    \multicolumn{1}{c|}{FreeKD-Prompt}  & GCN  & \multicolumn{1}{c|}{GAT}  & 87.15(\textbf{$\uparrow$2.03}) & \multicolumn{1}{c|}{87.12(\textbf{$\uparrow$1.67})} & 38.75(\textbf{$\uparrow$5.66})  & \multicolumn{1}{c|}{45.55(\textbf{$\uparrow$5.26})} & 78.56(\textbf{$\uparrow$3.14}) & \multicolumn{1}{c|}{77.75(\textbf{$\uparrow$2.09})} & 67.84(\textbf{$\uparrow$10.27})      & 65.95(\textbf{$\uparrow$8.11})      \\
    \multicolumn{1}{c|}{FreeKD-Prompt}  & GCN  & \multicolumn{1}{c|}{SAGE}  & 87.25(\textbf{$\uparrow$2.13}) & \multicolumn{1}{c|}{87.48(\textbf{$\uparrow$2.12})} & 38.07(\textbf{$\uparrow$4.98})  & \multicolumn{1}{c|}{50.50(\textbf{$\uparrow$1.73})} & 78.68(\textbf{$\uparrow$3.26}) & \multicolumn{1}{c|}{78.38(\textbf{$\uparrow$1.82})} & 66.76(\textbf{$\uparrow$9.19})      & 79.46(\textbf{$\uparrow$3.24})      \\
    \multicolumn{1}{c|}{FreeKD-Prompt}  & GAT  & \multicolumn{1}{c|}{SAGE}  & 87.15(\textbf{$\uparrow$1.70}) & \multicolumn{1}{c|}{87.45(\textbf{$\uparrow$2.09})} & 44.47(\textbf{$\uparrow$4.18})  & \multicolumn{1}{c|}{50.20(\textbf{$\uparrow$1.43})} & 77.97(\textbf{$\uparrow$2.31}) & \multicolumn{1}{c|}{78.31(\textbf{$\uparrow$1.75})} & 65.41(\textbf{$\uparrow$7.57})      & 78.65(\textbf{$\uparrow$2.43})      \\ 
    \bottomrule
    \end{tabular}%
    }
  \vspace{-0.2in}
  \label{trasductive}
\end{table*}

\begin{table}
  \centering
  \caption{Results (\%) of the compared approaches for node classification in the inductive setting on the PPI dataset.
  }
  \vspace{-0.1in}
  \renewcommand\arraystretch{1.1}
  \setlength{\tabcolsep}{1.2mm}{%
  \begin{tabular}{@{}ccccc@{}}
  \toprule
                              & \multicolumn{2}{c}{}             & \multicolumn{2}{c}{PPI}               \\ \midrule
  \multicolumn{1}{c|}{Method} & \multicolumn{2}{c|}{Basic Model}     & \multicolumn{2}{c}{F1 Score ($\uparrow$Impv.)} \\
  \multicolumn{1}{c|}{}       & $\Phi$ & \multicolumn{1}{c|}{$\Psi$} & $\Phi$              & $\Psi$              \\
  \midrule
  \multicolumn{1}{c|}{SAGE} & - & \multicolumn{1}{c|}{-}     & 69.28             &       -            \\
  \multicolumn{1}{c|}{GAT} & -  & \multicolumn{1}{c|}{-}     & 97.30              &        -           \\
  \midrule
  \multicolumn{1}{c|}{FreeKD}  & SAGE & \multicolumn{1}{c|}{SAGE} & 71.72(\textbf{$\uparrow$2.44})      & 71.56(\textbf{$\uparrow$2.28})      \\
  \multicolumn{1}{c|}{FreeKD}  & GAT  & \multicolumn{1}{c|}{GAT}  & 98.79(\textbf{$\uparrow$1.49})      & 98.73(\textbf{$\uparrow$1.43})      \\
  \multicolumn{1}{c|}{FreeKD}  & GAT  & \multicolumn{1}{c|}{SAGE} & 98.61(\textbf{$\uparrow$1.31})      & 72.39(\textbf{$\uparrow$3.11})      \\
  \midrule
  \multicolumn{1}{c|}{FreeKD-Prompt}  & SAGE & \multicolumn{1}{c|}{SAGE} & 72.95(\textbf{$\uparrow$3.67})      & 72.92(\textbf{$\uparrow$3.64})      \\
  \multicolumn{1}{c|}{FreeKD-Prompt}  & GAT  & \multicolumn{1}{c|}{GAT} & 98.96(\textbf{$\uparrow$1.66})      & 98.96(\textbf{$\uparrow$1.66})      \\
  \multicolumn{1}{c|}{FreeKD-Prompt}  & GAT  & \multicolumn{1}{c|}{SAGE} & 98.81(\textbf{$\uparrow$1.51})      & 74.28(\textbf{$\uparrow$5.00})      \\
  \bottomrule
  \end{tabular}%
  }
  \vspace{-0.2in}
  \label{inductive}
\end{table}

\subsubsection{Implementation Details}
\label{details}
 All the results are averaged over 10 times and we run our experiments on GeForce RTX 2080 Ti GPU.
 We use the Adam optimizer \cite{kingma2014adam} for training and adopt early stopping with a patience on validation sets of $150$ epochs. The initial learning rate is $0.05$ for GAT and  $0.01$ for GCN, GraphSAGE, and is decreased by multiplying $0.1$ every $100$ epochs. For the reinforced knowledge judge module, we set a fixed learning rate of $0.01$. We set the dropout rate to $0.5$ and the $l_2$ norm regularization weight decay to $0.0005$. The parameters of all GNN models are randomly initialized.  
The hyper-parameters  $\mu$ and $\rho$ in our method are searched from $\{0.5, 1.0,1.5,2.0\}$, and $\gamma$ is searched from $\{0.1,0.3,0.5\}$. The node-level policy network and structure-level policy network are both 3-layer MLP with tanh activation function and the size of hidden layer is set to $\{64,32\}$.
For the transductive setting, the number of layers in GNNs is set to $2$ and
the hidden size is set to $64$. For the inductive setting, the number
of layers in GNNs is set to $3$ and the hidden size is set to $256$. For GAT, the attention dropout probability is set to $0.5$ and the number of attention heads is set to $8$. For GraphSAGE, we use the mean aggregator to sample neighbors.

For FreeKD-Prompt, we configure the number of prompt tokens $P$, to be $100$ for Cora, Citeseer, Texas, and $50$ for Chameleon, PPI. We employ $M=5$ prompt graphs on all datasets. Regarding the percentage threshold $T_{\varphi_c}$ that determines edges between prompt nodes and input graph nodes, we set it at $0.5\%$ for Cora, Citeseer, and Chameleon, and $20\%$ for Texas. This means that there exists $1354$, $1663$, $1138$, $3360$, $14236$ edges between  prompt nodes and input graph nodes on Cora, Citeseer, Chameleon, Texas, PPI respectively. 
Additionally, the percentage threshold for edges between prompt nodes, $T_{\varphi_p}$, is uniformly set at $5\%$ across all datasets. 

In the experiments of comparison with other knowledge distillation methods, the student model, i.e., GAT, is set to 2-layer with 64 hidden size in the transductive setting and 3-layer with 256 hidden size in the inductive setting.
Here we adopt a typical GNN model GCNII \cite{chen2020simple} as the teacher model for other knowledge distillation methods.
For the teacher model GCNII, the number of layers is set to $32$ and the hidden size is set to $128$ in the transductive setting; in the inductive setting, the number of layers is set to $9$ and the hidden size is set to $2048$.
For all the compared knowledge distillation methods, we use the parameters as their original papers suggest and report their best results.

\subsection{Overall Evaluations on Our Method}
In this subsection, we evaluate our methods using three popular GNN models, GCN \cite{kipf2016semi}, GAT \cite{velivckovic2017graph}, and GraphSAGE \cite{hamilton2017inductive}. 
We arbitrarily select two networks from the above three models as our basic models $\Phi$ and $\Psi$, and perform our proposed FreeKD and FreeKD-Prompt, enabling them to learn from each other. 
Note that we do not perform GCN on the PPI dataset, because of the inductive setting.

	
Table \ref{trasductive} and Table \ref{inductive} report the experimental results.
As shown in Table \ref{trasductive} and Table \ref{inductive}, our FreeKD can consistently promote the performance of the basic GNN models in a large margin on all the datasets.
For instance, our FreeKD can achieve more than $4.5\%$ improvement by mutually learning from two GCN models on the Chameleon dataset, compared with the single GCN model. In summary, for the transductive learning tasks,  our FreeKD improves the performance by 1.01\% $\sim$ 1.97\% on the Cora and Citeseer datasets and 1.01\% $\sim$ 4.70\% on the Chameleon and Texas datasets, compared with the corresponding GNN models. 
For the inductive learning task, our FreeKD improves the performance by 1.31\% $\sim$ 3.11\% on the PPI dataset dataset.

After further employing FreeKD-Prompt, we observe a significant improvement in the performance of the GNN models.
Notably, FreeKD-Prompt can yield up to $2.34\%$, $6.98\%$, $3.26\%$ $10.27\%$, $5.00\%$ improvements on the Cora, Chameleon, Citeseer, Texas, PPI datasets, respectively.
This compelling improvement highlights the effectiveness of learning adaptive prompt-based augmentations in facilitating the transfer of diverse knowledge.
In addition, we observe that two GNN models either sharing the same architecture or using different architectures can both benefit from each other by using our FreeKD and FreeKD-Prompt, which shows the efficacy to various GNN models.
\vspace{-0.1in}
\subsection{Comparison with Knowledge Distillation}

Since our method is related to knowledge distillation, we also compare with the existing knowledge distillation methods to further verify effectiveness of our method. In this experiment, we first compare with three traditional knowledge distillation methods, KD \cite{hinton2015distilling}, LSP \cite{yang2020distilling}, CPF \cite{yang2021extract}, G-CRD \cite{joshi2022representation} distilling knowledge from a deeper and stronger  teacher GCNII model \cite{chen2020simple} into a shallower student GAT model. 
The structure details of GCNII and GAT could be found in Section \ref{details}.
In addition, we also compare with an ensemble learning method, RDD \cite{zhang2020reliable}, where a complex teacher network is generated by ensemble learning for distilling knowledge. Finally, we take GNN-SD \cite{chen2020self} as another baseline, which distills knowledge  from shallow layers into deep layers in one GNN.
For our method, we take two GAT sharing the same structure as the basic models.

Table \ref{kd} lists the experimental results. Surprisingly, our FreeKD perform comparably or even better than the traditional knowledge distillation methods (KD, LSP, CPF, G-CRD) on all the datasets. This demonstrates the effectiveness of our method, as they  distill knowledge from the stronger teacher GCNII while we only mutually distill knowledge between two shallower GAT.
In addition, our FreeKD consistently outperforms GNN-SD and RDD, which further illustrates the effectiveness of our proposed  FreeKD.
Finally, our FreeKD-Prompt obtains better performance than all baselines, indicating the effectiveness of our proposed prompt-enhanced knowledge distillation approach.

\begin{table}
\centering
\caption{Results (\%) of  different knowledge distillation methods. '-' means not available.
}
\vspace{-0.1in}
\renewcommand\arraystretch{1.1}
\setlength{\tabcolsep}{2mm}{
\begin{tabular}{c|c|c|c|c|c}
 \toprule
        & Cora                  & Chameleon              & Citeseer              & Texas                  & PPI                    \\ \midrule
Teacher & \multirow{2}{*}{87.80} & \multirow{2}{*}{46.84} & \multirow{2}{*}{78.60} & \multirow{2}{*}{65.14} & \multirow{2}{*}{99.41} \\
GCNII   &                       &                        &                       &                        &                        \\ \midrule
KD      & 86.13                 & 43.64                  & 77.03                 & 59.46                  & 97.81                  \\ 
LSP     & 86.25                 & 44.04                  & 77.21                 & 59.73                  & 98.25                  \\
CPF     & 86.41                 & 42.19                   & 77.80                  & 60.81                  & -                      \\ 
G-CRD    & 86.43              &   43.77                 &   77.62               & 60.54                  &  98.37                     \\  \midrule
GNN-SD  & 85.75                 & 40.92                  & 75.96                 & 58.65                  & 97.73                  \\
RDD     & 85.84                 & 41.73                  & 76.02                 & 58.92                  & 97.66                  \\ \midrule
FreeKD    & 86.68    & 44.32  & 77.42                 & 61.35      & 98.79  \\  
FreeKD-Prompt       & \textbf{87.44}    & \textbf{45.68}  & \textbf{78.05}                 & \textbf{64.87}      & \textbf{98.96}  \\  
\bottomrule             
\end{tabular}%
}
  \vspace{-0.2in}
\label{kd}
\end{table}

\begin{figure*}
\centering
\subfigure[$\Phi$: without noise; $\Psi$: without noise]{
\begin{minipage}[t]{0.33\linewidth}
\centering
\includegraphics[width=2in]{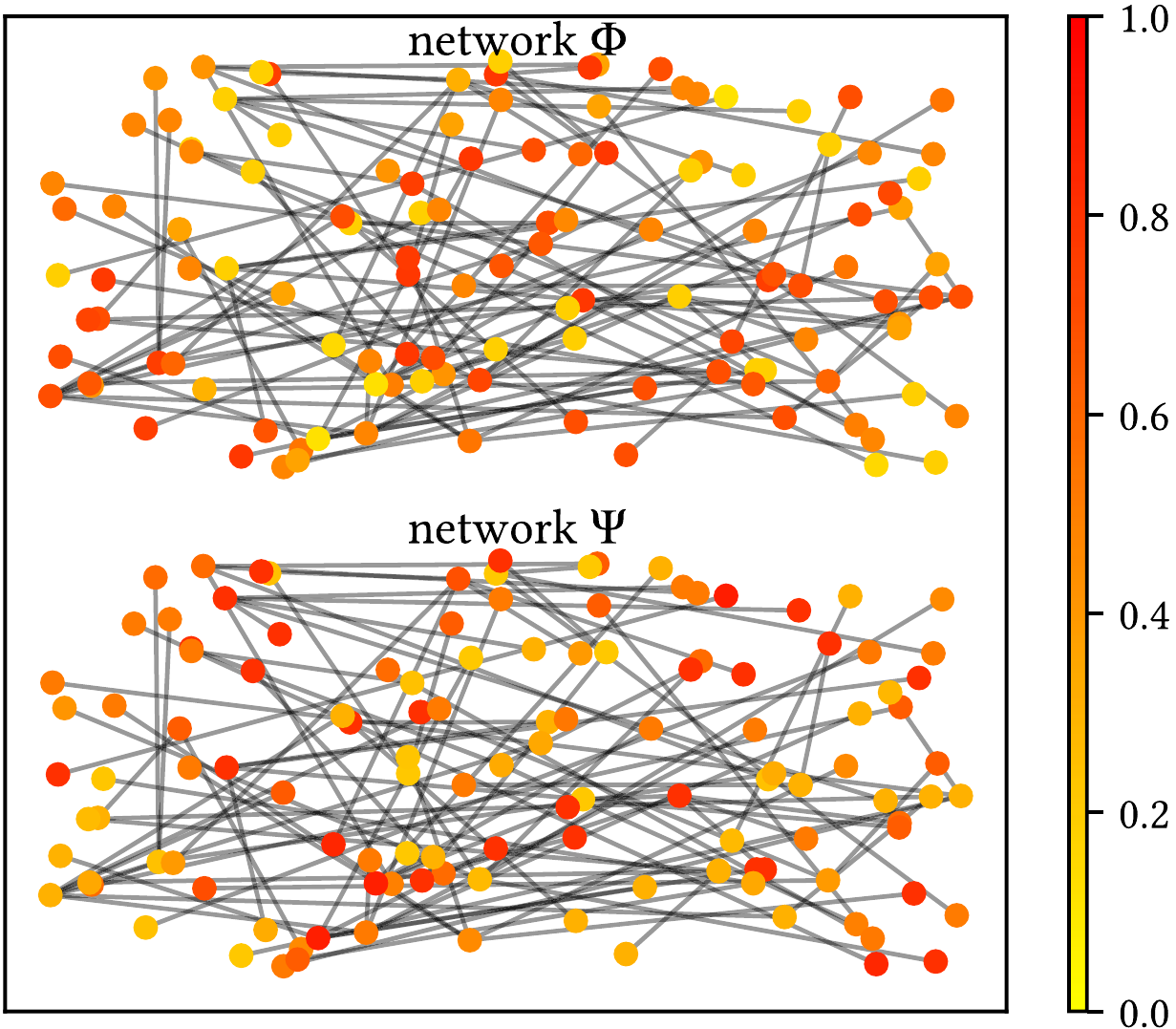}
\end{minipage}%
}%
\subfigure[$\Phi$: noise $\sigma=0.5$; $\Psi$: without noise]{
\begin{minipage}[t]{0.33\linewidth}
\centering
\includegraphics[width=2in]{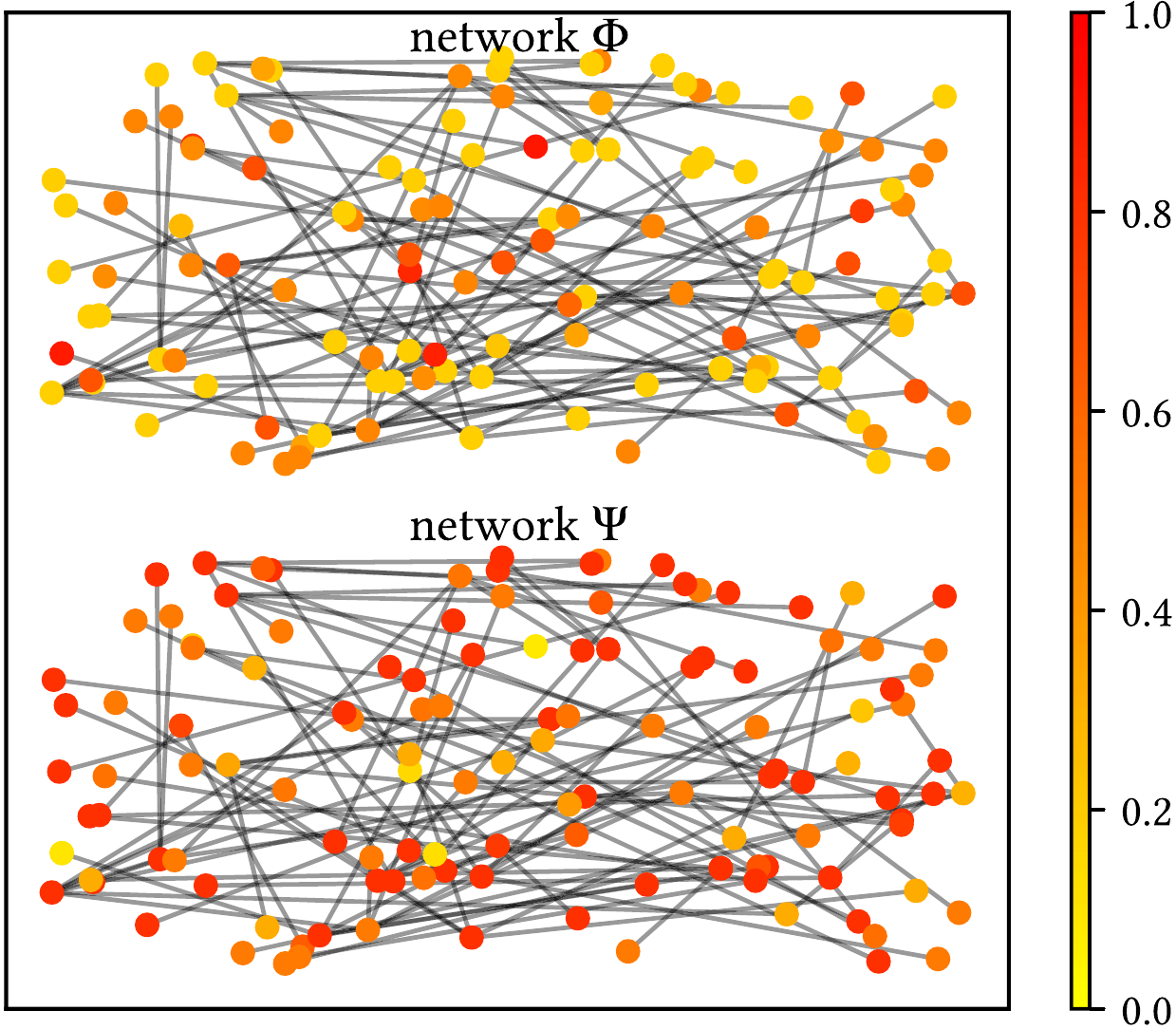}
\end{minipage}%
\label{vis2}
}%
\subfigure[$\Phi$: noise $\sigma=1.0$; $\Psi$: without noise]{
\begin{minipage}[t]{0.33\linewidth}
\centering
\includegraphics[width=2in]{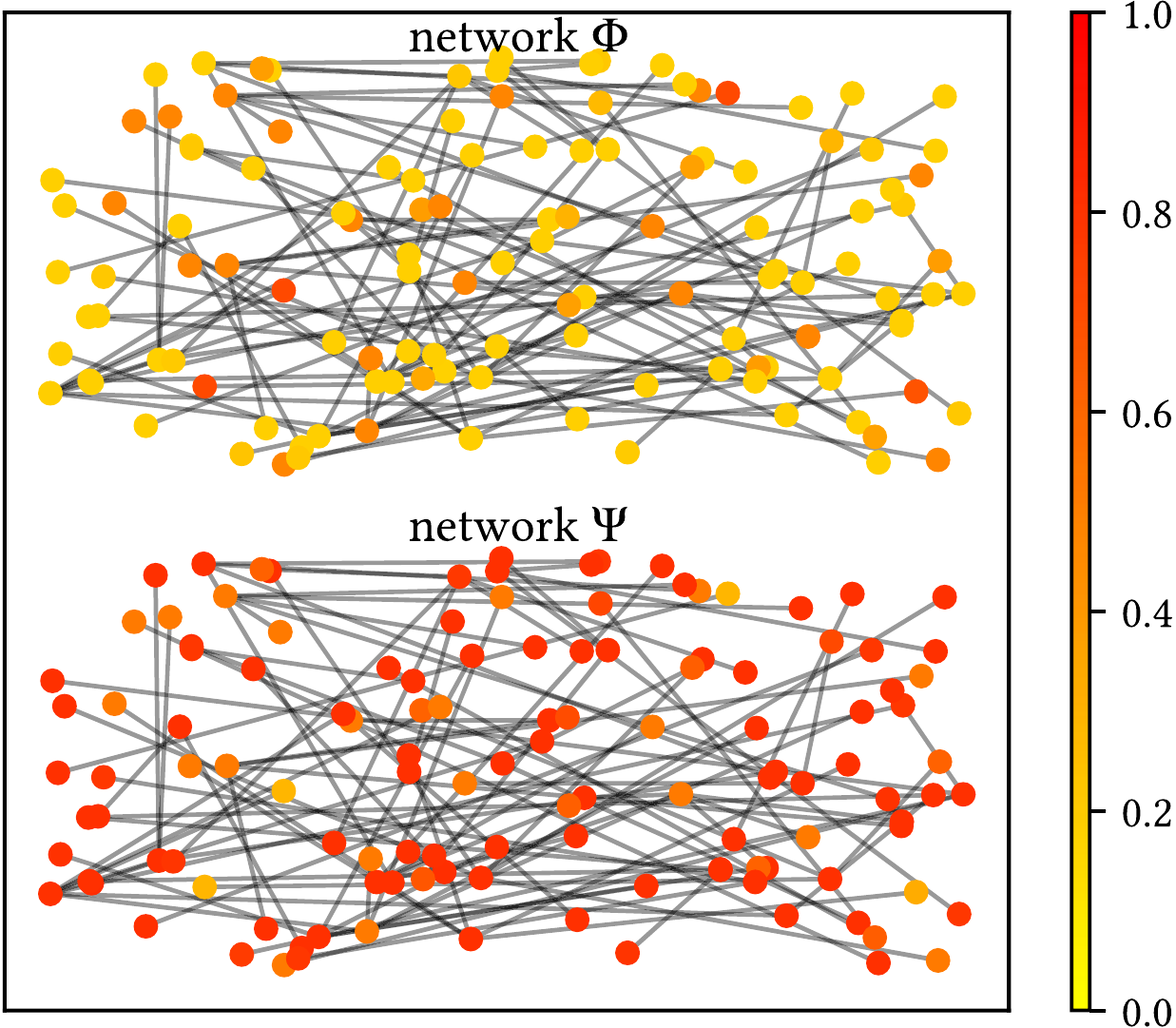}
\end{minipage}
\label{vis3}
}%
\centering
  \vspace{-0.1in}
  \caption{The output probabilities of our node-level policy reinforced knowledge judge module by adding different degrees of noise to the network $\Phi$. 
  The redder the node in a network is, the higher the probability for the node in this network to serve as a teacher node to transfer knowledge to the corresponding node of the other network is.}
  \label{vis}
  \vspace{-0.2in}
\end{figure*}

\begin{table}
  \centering
  \vspace{-0.1in}
  \caption{Ablation Study on the Cora and Chameleon dataset.}
  \vspace{-0.1in}
  \renewcommand\arraystretch{1.1}
   \setlength{\tabcolsep}{1mm}{
  \begin{tabular}{@{}ccccccc@{}}
  \toprule
  & & &\multicolumn{2}{c}{Cora} &\multicolumn{2}{c}{Chameleon} \\ 
  \midrule
  \multicolumn{1}{c|}{Method}                 & \multicolumn{2}{c|}{Network}     & \multicolumn{2}{c|}{F1 Score}     & \multicolumn{2}{c}{F1 Score}  \\
  \multicolumn{1}{c|}{}                       & $\Phi$ & \multicolumn{1}{c|}{$\Psi$} & $\Phi$  & \multicolumn{1}{c|}{$\Psi$}    & $\Phi$  & \multicolumn{1}{c}{$\Psi$}   \\
  \midrule
  \multicolumn{1}{c|}{GCN} & - & \multicolumn{1}{c|}{-} & 85.12 &\multicolumn{1}{c|}{-}  & 33.09 & - \\
  \multicolumn{1}{c|}{FreeKD-node} & GCN & \multicolumn{1}{c|}{GCN} & 86.17 & \multicolumn{1}{c|}{86.03}  &  36.18 &  36.16 \\
  \midrule
  \multicolumn{1}{c|}{FreeKD-w.o.-judge}     & GCN  & \multicolumn{1}{c|}{GCN}  & 85.83 & \multicolumn{1}{c|}{85.76}          &  35.48 &  35.09  \\
  \multicolumn{1}{c|}{FreeKD-loss} & GCN  & \multicolumn{1}{c|}{GCN}  & 85.89 & \multicolumn{1}{c|}{85.97}           &  35.77 &  36.10 \\
  \multicolumn{1}{c|}{FreeKD-all-neighbors} & GCN  & \multicolumn{1}{c|}{GCN}  & 86.21 & \multicolumn{1}{c|}{86.26}      &  36.54 & 36.27 \\
  \multicolumn{1}{c|}{FreeKD-all-structures} & GCN  & \multicolumn{1}{c|}{GCN}  & 86.13 & \multicolumn{1}{c|}{86.07}     &  36.47 & 36.49   \\
  \multicolumn{1}{c|}{FreeKD}                  & GCN  & \multicolumn{1}{c|}{GCN}  & 86.53 & \multicolumn{1}{c|}{86.62}    &   37.48 & 37.79 \\ \midrule
  \multicolumn{1}{c|}{FreeKD-Prompt-w.o.-info}  & GCN  & \multicolumn{1}{c|}{GCN}     & 86.74 & \multicolumn{1}{c|}{86.93}      &  37.91 & 38.11 \\ 
  \multicolumn{1}{c|}{FreeKD-Prompt-w.o.-div}   & GCN  & \multicolumn{1}{c|}{GCN}     & 87.16 & \multicolumn{1}{c|}{86.97}      &  39.30 & 39.25 \\ 
  \multicolumn{1}{c|}{FreeKD-Prompt}            & GCN  & \multicolumn{1}{c|}{GCN}     & 87.33 & \multicolumn{1}{c|}{87.46}      &  40.04 & 40.07 \\ 
  \bottomrule
  \end{tabular}
  }
   \label{ablation_study}
\end{table}

\subsection{Ablation Study}
We perform ablation study to verify the effectiveness of the components in our method. We use GCN as the basic models $\Phi$ and $\Psi$ in our method, and conduct the experiments on two datasets of different domains, Chameleon and Cora. 
When setting $\rho=0$, this means that we only transfer the node-level knowledge. We denote it FreeKD-node for short.
To evaluate our reinforcement learning based node judge module, we design three variants:

\begin{itemize}
    \item FreeKD-w.o.-judge: our FreeKD  without using the agent.
     $\Phi$ and $\Psi$ distills knowledge for each node from each other.
    \item FreeKD-loss: our FreeKD  without using the reinforced knowledge judge. It determines the directions of knowledge distillation only relying on the cross entropy loss.
    \item FreeKD-all-neighbors: our FreeKD selecting the directions of node-level knowledge distillation via node-level actions, but using all neighborhood nodes as the local structure.
    \item FreeKD-all-structures: our FreeKD selecting the directions of node-level knowledge distillation, but without using structure-level actions for  structure-level knowledge distillation.
\end{itemize}

Table \ref{ablation_study} shows the  results. 
FreeKD-node is better than GCN, showing that mutually transferring node-level knowledge via reinforcement learning is useful for boosting the performance of GNNs.
FreeKD obtain better results than FreeKD-node. It illustrates distilling structure knowledge by our method is beneficial to GNNs.
FreeKD achieves better performance than FreeKD-w.o.-judge, illustrating dynamically determining the knowledge distillation direction is important.
In addition, FreeKD outperforms FreeKD-loss. This shows that directly using the cross entropy loss to decide the directions of knowledge distillation is sub-optimal. As stated before, this heuristic strategy only considers the performance of the node itself, but neglects the influence of the node on other nodes.
Additionally, FreeKD has superiority over FreeKD-all-neighbors, demonstrating that transferring part of neighborhood information selected by our method is more effective than transferring all neighborhood information for GNNs. 
Finally, FreeKD obtains better performance than FreeKD-all-structures, which indicates our reinforcement learning based method can transfer more reliable structure-level knowledge. 

Furthermore, we conduct ablation study to validate the effectiveness of each loss function in FreeKD-Prompt. We design two variants of FreeKD-Prompt:
\begin{itemize}
    \item FreeKD-Prompt-w.o.-info: our FreeKD-Prompt without using the information preserving loss to learn the prompts.
    \item FreeKD-Prompt-w.o.-div: our FreeKD-Prompt without using the diversity loss to learn the prompts.
\end{itemize}
As illustrated in Table \ref{ablation_study}, our FreeKD-Prompt demonstrates superior performance compared to FreeKD-Prompt-w.o.-info and FreeKD-Prompt-w.o.-div. This highlights the effectiveness of the two loss functions employed in FreeKD-Prompt. In summary, these results demonstrate our proposed knowledge distillation framework is effective.

\begin{figure*}
\centering
\subfigure[Cora]{
\begin{minipage}[t]{0.24\linewidth}
\centering
\includegraphics[width=1.65in]{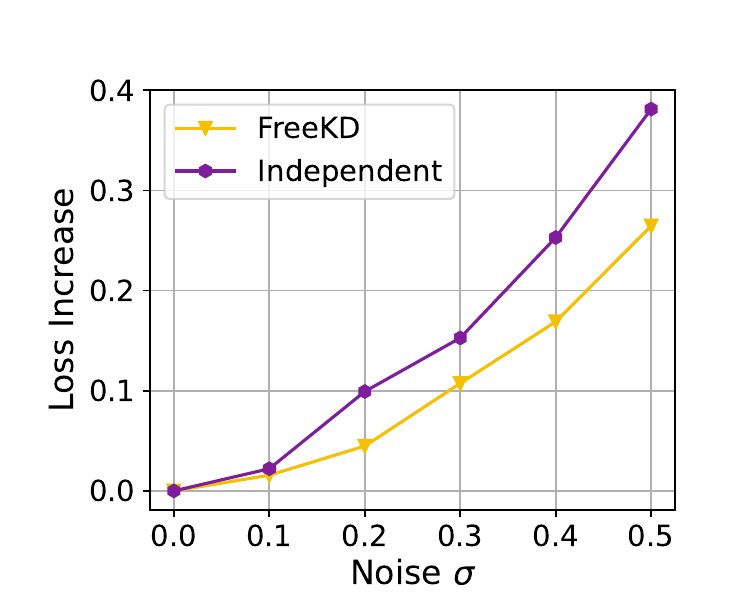}
\end{minipage}%
}%
\subfigure[Chameleon]{
\begin{minipage}[t]{0.24\linewidth}
\centering
\includegraphics[width=1.65in]{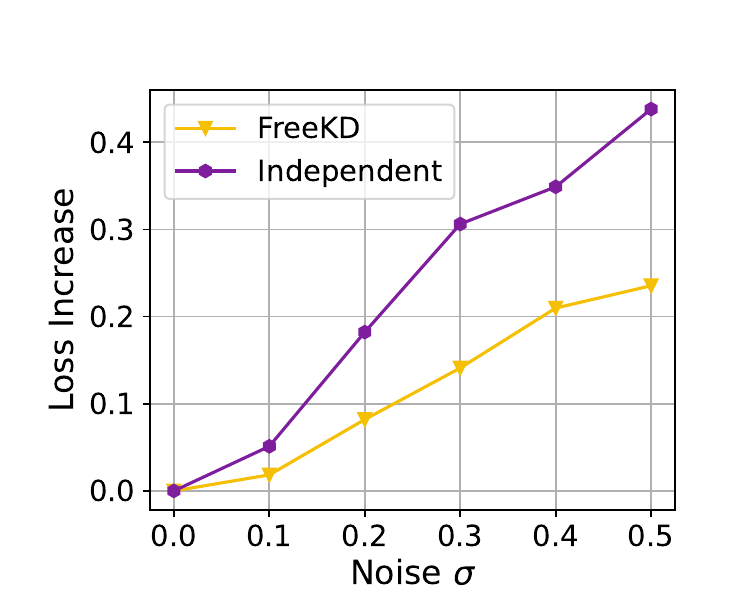}
\end{minipage}%
}%
\subfigure[Citeseer]{
\begin{minipage}[t]{0.24\linewidth}
\centering
\includegraphics[width=1.65in]{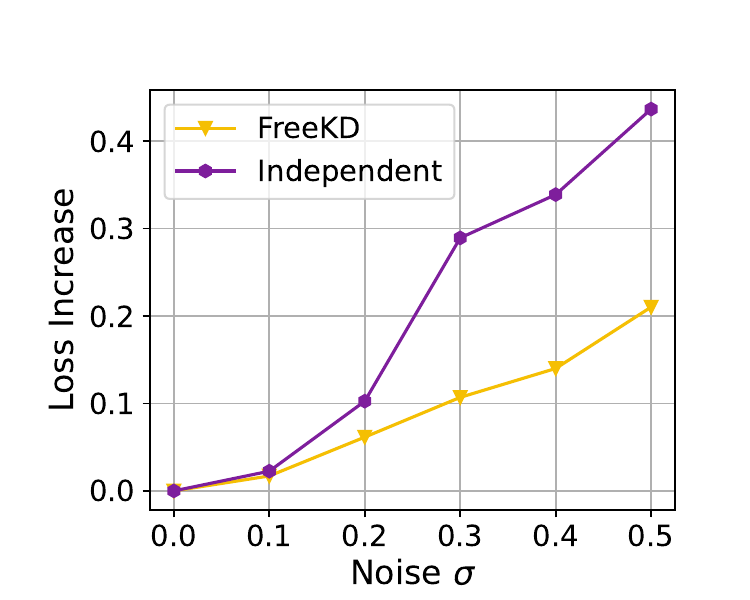}
\end{minipage}%
}%
\subfigure[Texas]{
\begin{minipage}[t]{0.24\linewidth}
\centering
\includegraphics[width=1.67in]{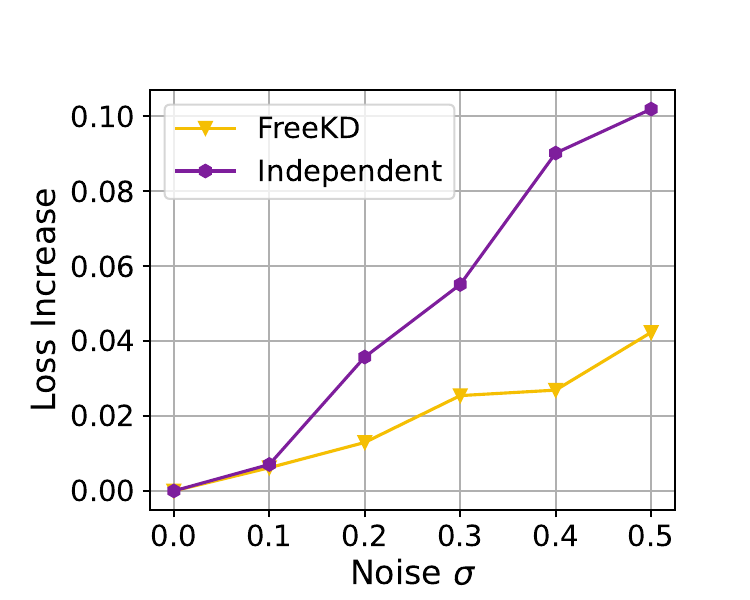}
\end{minipage}%
}%
\centering
\vspace{-0.1in}
\caption{Loss increase by varying levels of noise to GCN trained by FreeKD and independently, respectively.}
\label{loss_change}
\vspace{-0.2in}
\end{figure*}

\begin{figure}
  \centering
  \includegraphics[width=0.9\linewidth]{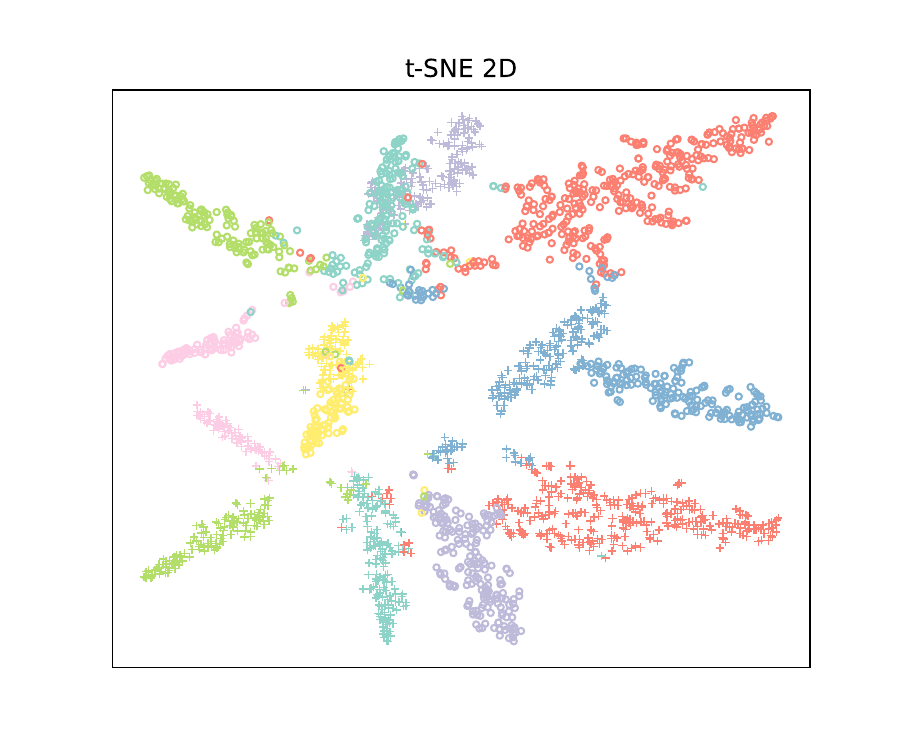}
   \vspace{-0.1in}
  \caption{
    t-SNE visualizations \cite{van2008visualizing} of output features in the last layer of two GCNs trained by FreeKD. The different color denotes different classes and  different shapes ('+' and 'o') denotes different GCNs, respectively.
  }
  \label{vis_cora}
  \vspace{-0.2in}
\end{figure}

\subsection{Comparison with Graph Augmentation}
Since our proposed FreeKD-Prompt has connections with graph augmentation, we also compare our FreeKD-Prompt with two widely used graph augmentation methods (DropNode \cite{rong2019dropedge}, DropEdge \cite{feng2020graph}) and a recent learning-based graph augmentation approach (HalfHop \cite{azabou2023half}). 
For fair comparison, we utilize these methods to generate multiple graph augmentations and adopt these augmentations to train the GNNs with FreeKD. In this way, we derive three variants of our method: FreeKD+DropNode, FreeKD+DropEdge and FreeKD+HalfHop. Besides, we also design a hybrid baseline FreeKD+DN+DE+HH by integrating these three methods together into FreeKD. 
The results are illustrated in Table \ref{aug_study}, it's evident that our FreeKD-Prompt outperforms these methods by a large margin. This demonstrates the superiority of learning undistorted and diverse graph augmentations based on prompt learning for distilling distinct knowledge.

\subsection{Visualizations}
We further intuitively show the effectiveness of the reinforced knowledge judge to dynamically decide the directions of knowledge distillation.
We set GCN as $\Phi$ and GraphSAGE as $\Psi$, and train our FreeKD on the Cora dataset.
Then, we poison  $\Phi$ by adding random Gaussian noise with a standard deviation $\sigma$ to its model parameters. Finally, we visualize the agent’s output, i.e., node-level policy probabilities $\pi_\mathbf{\theta}\left(\mathbf{s}_i^{[1]},0\right)$  and $\pi_\mathbf{\theta}\left(\mathbf{s}_i^{[1]},1\right)$ at node $i$ for $\Phi$ and $\Psi$, respectively. To better visualize, we show a subgraph composed of the first 30 nodes and their neighborhoods.

Fig. \ref{vis} shows the results using different standard deviations $\sigma$. 
In Fig. \ref{vis} (a), (b), and (c), the higher the probability output by the agent is, the redder the node is. And this means that the probability for the node in this network to serve as a distilled node to transfer knowledge to the corresponding node of the other network is higher.
As shown in Fig. \ref{vis} (a), when without adding noise, the degrees of the red color in $\Phi$ and $\Psi$ are comparable. As the noise is gradually increased in $\Phi$, the red color becomes more and more light in $\Phi$, but an opposite case happens in $\Psi$, as shown in Fig. \ref{vis} (b) and (c). This is because the noise brings negative influence on the outputs of the network, leading to inaccurate soft labels and large losses. In such a case, our agent can output low probabilities for the network $\Phi$.  Thus, our agent can effectively determine the direction of knowledge distillation for each node.

\begin{table}
  \centering
  \caption{Comparison with different graph augmentation methods.}
  \vspace{-0.1in}
  \renewcommand\arraystretch{1.1}
   \setlength{\tabcolsep}{1.2mm}{
  \begin{tabular}{@{}ccccccc@{}}
  \toprule
  & & &\multicolumn{2}{c}{Cora} &\multicolumn{2}{c}{Chameleon} \\ 
  \midrule
  \multicolumn{1}{c|}{Method}                 & \multicolumn{2}{c|}{Network}     & \multicolumn{2}{c|}{F1 Score}     & \multicolumn{2}{c}{F1 Score}  \\
  \multicolumn{1}{c|}{}                       & $\Phi$ & \multicolumn{1}{c|}{$\Psi$} & $\Phi$  & \multicolumn{1}{c|}{$\Psi$}    & $\Phi$  & \multicolumn{1}{c}{$\Psi$}   \\
  \midrule
  \multicolumn{1}{c|}{FreeKD+DropNode}     & GCN  & \multicolumn{1}{c|}{GCN}  & 86.26 & \multicolumn{1}{c|}{85.94}          &  38.75 &  38.50  \\
  \multicolumn{1}{c|}{FreeKD+DropEdge} & GCN  & \multicolumn{1}{c|}{GCN}  & 86.30 & \multicolumn{1}{c|}{86.45}          &  38.86 &  38.70  \\
  \multicolumn{1}{c|}{FreeKD+HalfHop} & GCN  & \multicolumn{1}{c|}{GCN}  & 86.03 & \multicolumn{1}{c|}{86.28}          &  39.03 &  39.38  \\
  \multicolumn{1}{c|}{FreeKD+DN+DE+HH} & GCN  & \multicolumn{1}{c|}{GCN}  & 86.73 & \multicolumn{1}{c|}{86.88}          &  39.23 &  38.99  \\
  \multicolumn{1}{c|}{FreeKD-Prompt}         & GCN  & \multicolumn{1}{c|}{GCN}  & 87.33 & \multicolumn{1}{c|}{87.46}          &  40.04 &  40.07  \\ 
  \midrule
  \multicolumn{1}{c|}{FreeKD+DropNode}     & GAT  & \multicolumn{1}{c|}{GAT}  & 86.36 & \multicolumn{1}{c|}{86.28}          &  44.28 &  44.32  \\
  \multicolumn{1}{c|}{FreeKD+DropEdge} & GAT  & \multicolumn{1}{c|}{GAT}  & 86.07 & \multicolumn{1}{c|}{86.22}          &  44.32 &  44.96  \\
  \multicolumn{1}{c|}{FreeKD+HalfHop} & GAT  & \multicolumn{1}{c|}{GAT}  & 86.65 & \multicolumn{1}{c|}{86.57}          &  44.80 &  44.69  \\
  \multicolumn{1}{c|}{FreeKD+DN+DE+HH} & GAT  & \multicolumn{1}{c|}{GAT}  & 86.73 & \multicolumn{1}{c|}{86.82}          &  44.75 &  44.36  \\
  \multicolumn{1}{c|}{FreeKD-Prompt}         & GAT  & \multicolumn{1}{c|}{GAT}  & 87.44 & \multicolumn{1}{c|}{87.27}          &  45.68 &  45.35  \\ 
  \bottomrule
  
  \end{tabular}
  }
   \label{aug_study}
     \vspace{-0.2in}
\end{table}

\subsection{Further Understanding Why FreeKD is Effective}


In this subsection, we further investigate why our FreeKD is able to improve the performance of GNNs by asking the following two questions. (i) Does FreeKD help GNNs find a better minimum? (ii) Does FreeKD make the two GNNs so similar that there is no diverse knowledge to exchange, especially when they share the same architecture? We conduct experiments to answer these questions as follows.

\textbf{FreeKD helps GNNs find a more robust and flatter minimum.} 
Motivated by \cite{chaudharientropy}, we conduct experiments to investigate the robustness of the obtained minimum of the GNNs,  by adding random Gaussian noise with a standard deviation $\sigma$ to the model parameters of the GCN trained using FreeKD and independently, respectively. We then visualize the increase in the training loss under varying levels of noise, as depicted in Fig. \ref{loss_change}.
It can be found that the loss change of  GCN  trained using our FreeKD is considerably smaller compared to that trained independently. 
This indicates that our FreeKD can help GNN discover a more robust and flatter minimum, thus improving the generalization capability of the GNNs. 

One possible reason for this result might be that our FreeKD in some aspects has association with the regularization techniques (e.g. label smoothing \cite{muller2019does}, normalization techniques \cite{ba2016layer}). These regularization techniques are helpful for finding a more generalized minimum \cite{zheng2021regularizing,chaudharientropy}.
However, different from the conventional regularization techniques, our FreeKD regularizes the two GNNs towards a more 'reasonable' direction by making the output of each GNN match the 'beneficial' one in both node-level aspect and structure-level aspect.

\textbf{The two GNNs preserve diverse features during training by FreeKD, even if they share the same architecture.} We know that the two GNNs are optimized towards the same ground-truth label, and our FreeKD makes them learn from each other. In this case, an  issue is that the two GNNs might be too similar to effectively teach each other, especially when they share the same architecture. To investigate this, we visualize the output features in the last layer of two GCNs after training by FreeKD on Cora as shown in Fig. \ref{vis_cora}. The two GCNs share the same architecture and the only difference between them comes from the random initialization of model parameters. 
We can observe that after being trained by FreeKD, the two GCNs still exhibit distinct features.
This indicates that each GNN has its own strengths and weaknesses when processing graph data, allowing for the exchange of diverse knowledge throughout the training process, even if they share the same architecture.

\begin{table*}[]
  \centering
  \caption{Results (\%) of the FreeKD++ and FreeKD-Prompt++. 
  The performance of each GNN architecture records its mean performance across multiple (or single) networks.
  'AVG' denotes the average performance of the three GNN architectures. 'Vanilla' represents training the GNNs independently and 'FreeKD' represents training the GNNs by exchanging knowledge within two GNNs of the same architecture.}
  \vspace{-0.1in}
  \renewcommand\arraystretch{1.1}
  \setlength{\tabcolsep}{0.7mm}{%
  \begin{tabular}{cccc|cccc|cccc|cccc|cccc} \toprule
  \multirow{2}{*}{Method} & \multicolumn{3}{c}{\multirow{2}{*}{\#Network Count}}  & \multicolumn{4}{|c|}{Cora}        & \multicolumn{4}{c|}{Chameleon}   & \multicolumn{4}{c|}{Citeseer}    & \multicolumn{4}{c}{Texas}       \\ 
                          & \multicolumn{3}{c}{}                               &                         \multicolumn{4}{|c|}{F1 Score}    & \multicolumn{4}{c|}{F1 Score}    & \multicolumn{4}{c|}{F1 Score}    & \multicolumn{4}{c}{F1 Score}    \\ \midrule
                          & GCN          & SAGE          & GAT                        & GCN   & SAGE & GAT   & AVG & GCN   & SAGE & GAT   & AVG & GCN   & SAGE & GAT   & AVG & GCN   & SAGE & GAT   & AVG \\ \midrule
  Vanilla  & 1 & 1 & 1  & 85.12 & 85.36 & 85.45 & 85.31 & 33.09 & 48.77 & 40.29 & 40.72 & 75.42 & 76.56 & 75.66 & 75.88 & 57.57 & 76.22 & 57.84 & 63.88 \\ \midrule
  FreeKD   & 2 & 2 & 2  & 86.58 & 86.48 & 86.57 & 86.54 & 37.64 & 49.84 & 44.21 & 43.89 & 77.31 & 77.68 & 77.28 & 77.42 & 60.41 & 77.71 & 61.31 & 66.47 \\ \midrule
  FreeKD++ & 1 & 1 & 1  & 86.72 & 87.17 & 86.45 & 86.78 & 37.65 & 50.24 & 44.61 & 44.17 & 77.84 & 77.76 & 77.73 & 77.78 & 61.08 & 78.11 & 61.89 & 67.03 \\
  FreeKD++ & 2 & 2 & 2  & 86.60 & 86.86 & 87.36 & 86.94 & 37.75 & 50.40 & 44.39 & 44.18 & 77.90 & 77.71 & 78.02 & 77.88 & 61.62 & 78.51 & 61.49 & 67.21 \\
  FreeKD++ & 3 & 3 & 3  & 86.75 & 87.10 & 87.31 & 87.05 & 38.03 & 50.16 & 44.57 & 44.25 & 78.23 & 77.76 & 78.16 & 78.05 & 60.81 & 80.09 & 61.26 & 67.39 \\ 
  \midrule
  FreeKD-Prompt++ & 1 & 1 & 1 & 87.39 & 87.66 & 86.85 & 87.30 & 39.78 & 50.11 & 45.07 & 44.99 & 78.47 & 78.10 & 77.98 & 78.18 & 67.84 & 79.19 & 64.05 & 70.36 \\
  FreeKD-Prompt++ & 2 & 2 & 2 & 87.51 & 87.61 & 87.03 & 87.38 & 40.12 & 50.13 & 45.65 & 45.30 & 78.28 & 78.46 & 78.15 & 78.30 & 68.65 & 79.87 & 62.84 & 70.45 \\
  FreeKD-Prompt++ & 3 & 3 & 3 & 87.40 & 87.71 & 87.12 & 87.41 & 40.07 & 50.11 & 45.97 & 45.38 & 78.49 & 78.44 & 77.95 & 78.29 & 66.49 & 79.82 & 65.05 & 70.45 \\
  \bottomrule
  \end{tabular}
  }
  \label{view1}
  \vspace{-0.2in}
\end{table*}

\begin{figure}
\centering
\subfigure[GCN]{
\begin{minipage}[t]{0.49\linewidth}
\centering
\includegraphics[width=1.7in]{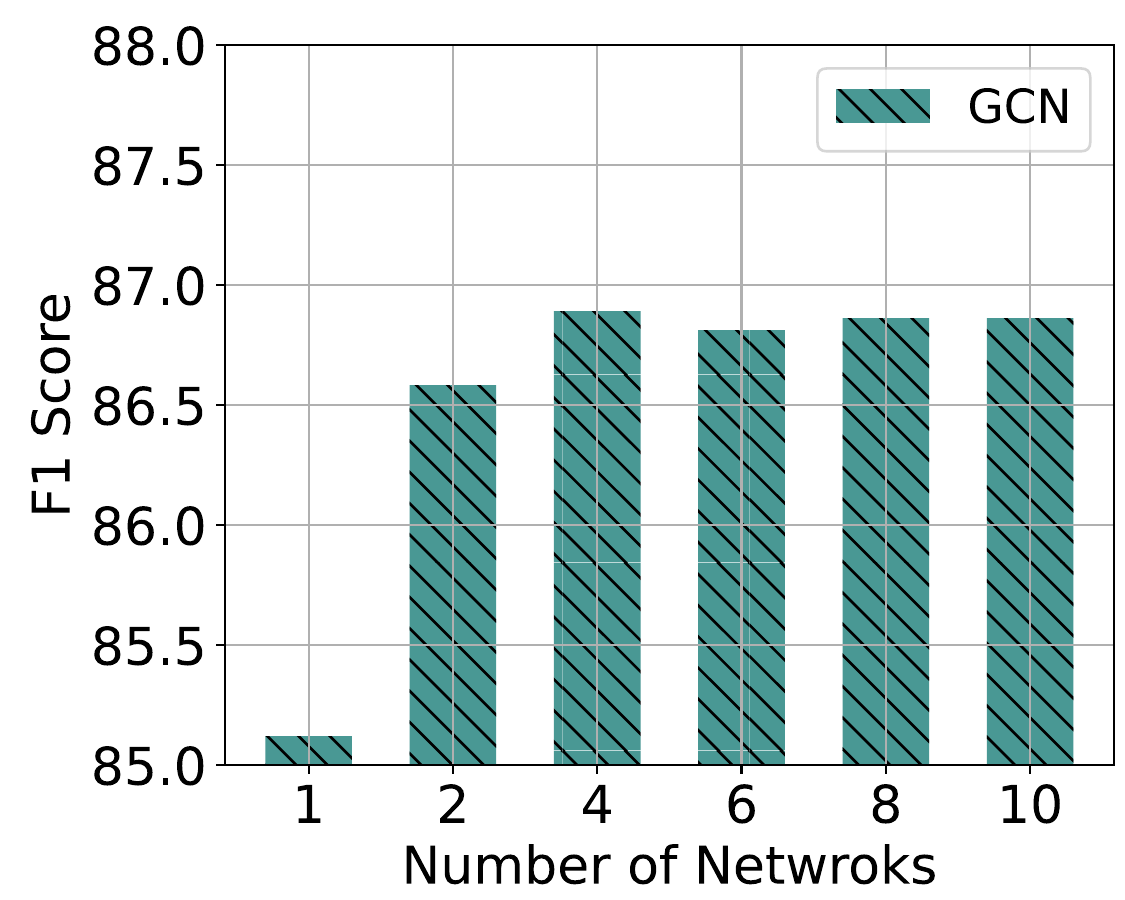}
\end{minipage}%
}%
\subfigure[GAT]{
\begin{minipage}[t]{0.49\linewidth}
\centering
\includegraphics[width=1.7in]{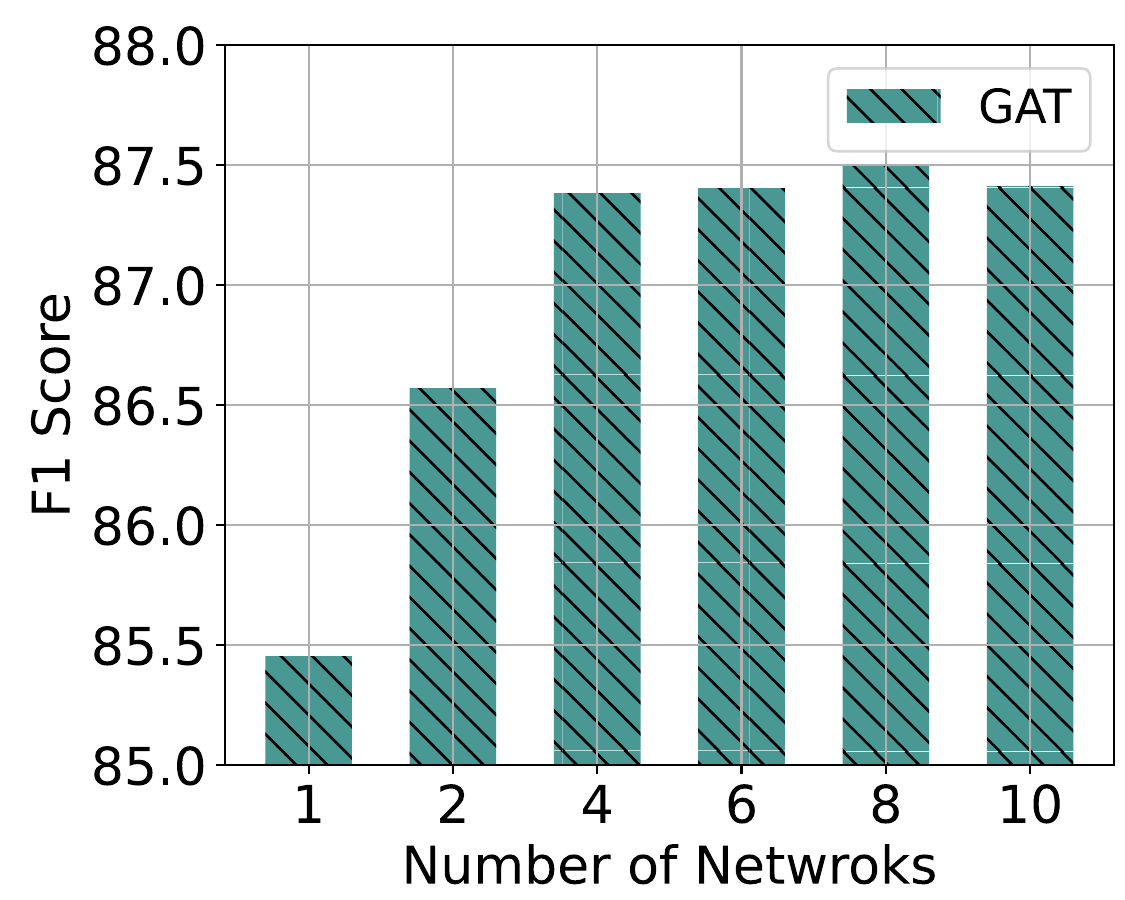}
\end{minipage}%
}%
\centering
\vspace{-0.1in}
\caption{Average Performance of multiple networks with different numbers of networks on the Cora dataset.}
\vspace{-0.2in}
\label{gnn_number}
\end{figure}

\subsection{Performance on Multiple GNNs}
We evaluate the performance of FreeKD++ and FreeKD-Prompt++ in this subsection, which extend our methods to multiple GNNs.

\textbf{Overall performance.} Table \ref{view1} presents the overall results of FreeKD++ and FreeKD-Prompt++. In this table, 'Vanilla' refers to training the GNNs independently and 'FreeKD' represents training the GNNs by exchanging knowledge between two GNNs of the same architecture. As for FreeKD++ and FreeKD-Prompt++, we exchange knowledge between multiple GNNs. 
From this table, we can find that FreeKD++ and FreeKD-Prompt++ generally outperform both the vanilla method and FreeKD. These results showcase the effectiveness of our methods in facilitating knowledge exchange between multiple GNNs.  By enabling such knowledge transfer, FreeKD++ and FreeKD-Prompt++ effectively enhances the overall performance of the GNN models.

\textbf{Effect of network number analysis.} Then, we investigate the effect of the number of networks in FreeKD++. 
We increase the number of GNN models of the same architecture, and plot the average performance of these models as shown in Fig. \ref{gnn_number}.
 We observe that  the performance of the model will be improved, as the number of networks increases. when the number of network  is larger than a certain threshold, the performance will become stable.

\subsection{Sensitivity and Convergence Analysis}

\begin{table}
\centering
\caption{Sensitivity study of $\gamma$ on the Cora dataset.}
\vspace{-0.1in}
\renewcommand\arraystretch{1.1}
\setlength{\tabcolsep}{0.9mm}{%
\begin{tabular}{@{}c|cc|cccccc@{}}
\toprule
\multirow{2}{*}{Dataset} & \multicolumn{2}{c|}{Network} & \multirow{2}{*}{$\gamma$=0.0} & \multirow{2}{*}{$\gamma$=0.1} & \multirow{2}{*}{$\gamma$=0.3} & \multirow{2}{*}{$\gamma$=0.5} & \multirow{2}{*}{$\gamma$=0.7} & \multirow{2}{*}{$\gamma$=0.9} \\
                         & $\Phi$          & $\Psi$         &                        &                        &                        &                        &                        &                        \\ \midrule

\multirow{2}{*}{Cora}    & GCN           & GCN          & 86.32                  & 86.39                  & \textbf{86.57}         & 86.31                  & 86.12                  & 85.23                   \\
                         & GAT           & GAT          & 86.21                  & 86.45                  & 86.41                  & \textbf{86.57}         & 86.32                  & 86.22                  \\ \bottomrule
\end{tabular}%
}
 \label{gamma}
 \vspace{-0.1in}
\end{table}

\begin{table}
  \centering
  \caption{Sensitivity study of $\beta$ on the Cora dataset.}
  \vspace{-0.1in}
  \renewcommand\arraystretch{1.1}
  \setlength{\tabcolsep}{0.9mm}{%
  \begin{tabular}{@{}c|cc|cccccc@{}}
  \toprule
  \multirow{2}{*}{Dataset} & \multicolumn{2}{c|}{Network} & \multirow{2}{*}{$\beta$=0.0} & \multirow{2}{*}{$\beta$=0.1} & \multirow{2}{*}{$\beta$=0.3} & \multirow{2}{*}{$\beta$=0.5} & \multirow{2}{*}{$\beta$=0.7} & \multirow{2}{*}{$\beta$=0.9} \\
                           & $\Phi$          & $\Psi$         &                        &                        &                        &                        &                        &                        \\ \midrule
  
  \multirow{2}{*}{Cora}    & GCN           & GCN & 87.07                & 87.11                & 87.14                & \textbf{87.40}       & 87.27                & 87.32                \\
                           & GAT           & GAT & 86.71                & 86.99                & 87.20                & \textbf{87.36}       & 87.22                & 87.20                \\ 
  \bottomrule
  \end{tabular}%
  }
   \label{beta}
   \vspace{-0.2in}
\end{table}

\begin{table}[t]
  \centering
  \caption{Results (\%) of FreeKD-Prompt with different numbers of prompt graphs $M$ on the Cora dataset.}
  \vspace{-0.1in}
  \renewcommand\arraystretch{1.1}
  \setlength{\tabcolsep}{0.9mm}{%
  \begin{tabular}{@{}c|cc|cccccc@{}}
  \toprule
  \multirow{2}{*}{Dataset} & \multicolumn{2}{c|}{Network}  & \multirow{2}{*}{$M$=2} & \multirow{2}{*}{$M$=3} & \multirow{2}{*}{$M$=4} & \multirow{2}{*}{$M$=5} & \multirow{2}{*}{$M$=6}  & \multirow{2}{*}{$M$=7} \\
                           & $\Phi$          & $\Psi$         &                        &                        &                        &                        &        &                                       \\ \midrule
  
  \multirow{2}{*}{Cora}    & GCN           & GCN & 87.19                & 87.26                & 87.33                & \textbf{87.40}       & 87.27                & 87.34                \\
                           & GAT           & GAT & 86.91                & 87.12                & 87.26                & 87.36                & \textbf{87.58}       & 87.32                \\ 
  \bottomrule
  \end{tabular}%
  }
   \label{param_m}
   \vspace{-0.2in}
\end{table}

\begin{table}[t]
  \centering
  \caption{Results (\%) of FreeKD-Prompt with different numbers of prompt tokens $P$ and varying link percentages $T_{\phi_c}$ on the Cora dataset.}
  \vspace{-0.1in}
  \renewcommand\arraystretch{1.1}
  \setlength{\tabcolsep}{1.2mm}{%
  \begin{tabular}{@{}c|cc|c|ccccc@{}}
  \toprule
  \multirow{2}{*}{Dataset} & \multicolumn{2}{c|}{Network} & \multirow{2}{*}{$T_{\phi_c}$} & \multirow{2}{*}{$P$=10} & \multirow{2}{*}{$P$=20} & \multirow{2}{*}{$P$=30} & \multirow{2}{*}{$P$=50}  & \multirow{2}{*}{$P$=100} \\ 
                           & $\Phi$          & $\Psi$         & & & & & &                   \\ \midrule
  
  \multirow{2}{*}{Cora}    & GCN           & GCN         & \multirow{2}{*}{0.5\%}    & 86.68 & 86.82          & 87.04          & 87.23          & \textbf{87.40} \\
                           & GAT           & GAT & & 87.17 & 87.31          & 87.25          & \textbf{87.41} & 87.36          \\ \midrule
  \multirow{2}{*}{Cora}    & GCN           & GCN         & \multirow{2}{*}{2\%}    & 86.83 & \textbf{87.15} & 87.13          & 87.09          & 86.47          \\
                           & GAT           & GAT & & 87.31 & 87.32          & \textbf{87.52} & 87.40          & 87.23         \\ \bottomrule
  \end{tabular}%
  }
   \label{param_p}
\end{table}

\begin{figure}
\centering
\subfigure[Sensitivity study of $\mu$ and $\rho$.]{
\begin{minipage}[t]{0.49\linewidth}
\centering
\includegraphics[width=1.6in]{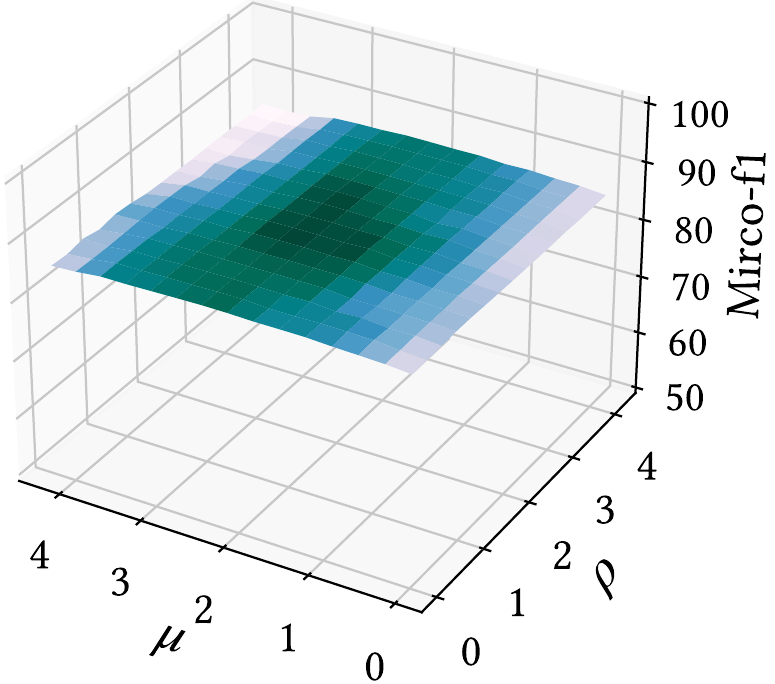}
\end{minipage}%
}%
\subfigure[Convergence curve.]{
\begin{minipage}[t]{0.49\linewidth}
\centering
\includegraphics[width=1.6in]{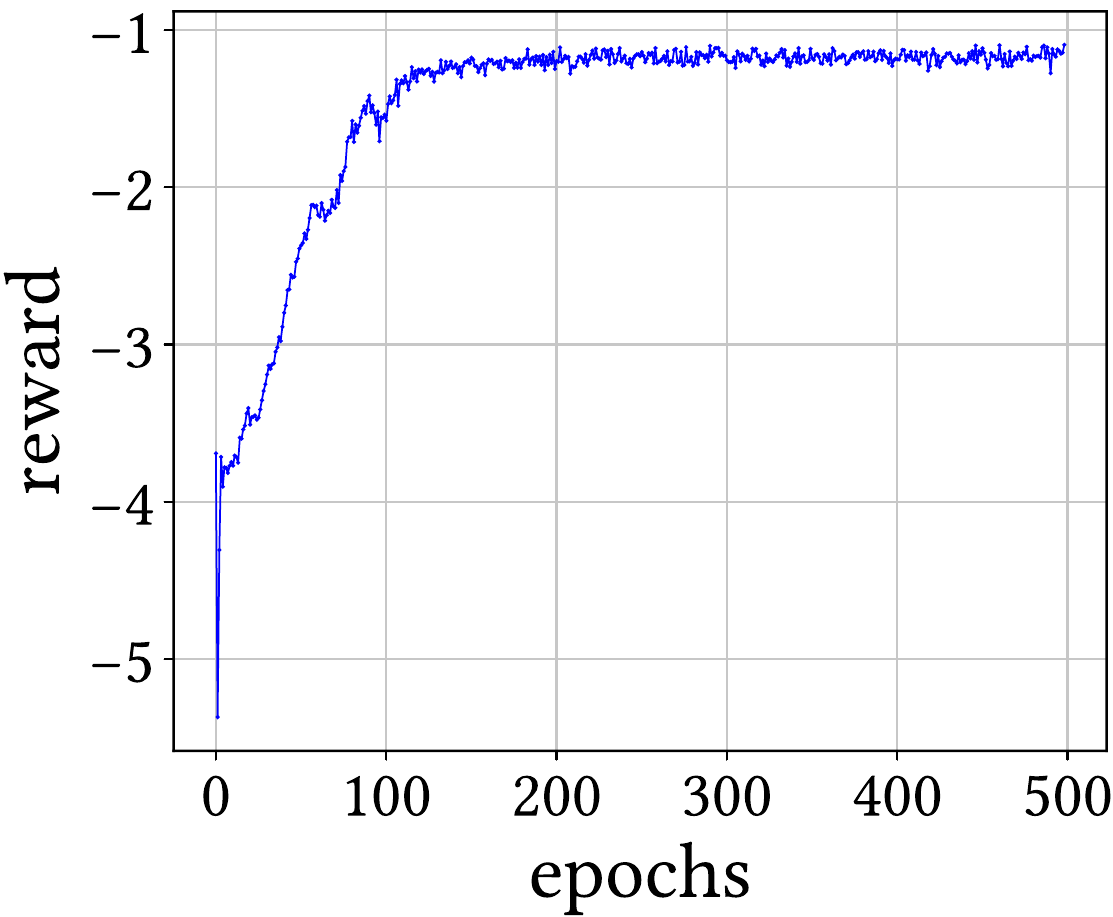}
\end{minipage}%
}%
\centering
\vspace{-0.1in}
\caption{Sensitivity study of $\mu$,$\rho$ and convergence analysis on the Cora dataset.}
\label{param}
\vspace{-0.2in}
\end{figure}

First, we analyze the sensitivity of three hyper-parameters in the loss function of FreeKD and one hyper-parameters in the prompt learning loss, i.e., $\gamma$ in the reward function (\ref{reward}),  $\mu$ and $\rho$ in the loss function (\ref{loss1}) and (\ref{loss2}), $\beta$ in the prompt learning loss function (\ref{ploss}). We study the sensitivity of our method to these hyper-parameters on the Cora dataset.
First, we investigate the impact of $\gamma$ in the agent’s reward function on the performance of our method. 
As shown in Table \ref{gamma}, with the values of $\gamma$ increasing, 
the performance of our method will fall after rising. In the meantime, our method is not sensitive to $\gamma$ in a relatively large range.
Additionally, we analyze the influence of $\beta$ in the prompt learning as shown in Table \ref{beta}. We can also find that our method is not sensitive to $\beta$ in a relatively large range. 
What's more, we also study the parameter sensitiveness of our method to $\mu$ and $\rho$. 
Fig. \ref{param}(a) shows the results. Our method is still not sensitive to these two hyper-parameters in a relatively large range.

What's more, we conduct experiments to explore the impact of hyper-parameters in the prompt graph design within FreeKD-Prompt.
First, we analyze the effect of the number of prompt tokens $P$ and the percentages $T_{\phi_c}$ for forming links between the prompt graph and the input graph. 
The results are shown in Table \ref{param_p}. We can find that our method is not sensitive to the number of prompt tokens within a reasonably wide range.
Moreover, when the number of prompt tokens is limited, augmenting the link percentage $T_{\phi_c}$ tends to enhance the performance.
Second, we analyze the effect of the number of prompt graphs $M$ as shown in Table \ref{param_m}. We observe that our method is also not sensitive to $M$ in a relatively large range.

Finally, we analyze the convergence of our method. Fig. \ref{param}(b) shows the reward convergence curve. It can be found that our method is convergent after around 100 epochs.

\section{Conclusion and Future Works}

In this paper, we propose a free-direction  knowledge distillation framework FreeKD to enable two shallower GNNs to learn from each other, without requiring  a deeper well-optimized teacher GNN. Meanwhile, we devise a hierarchical reinforcement learning  mechanism to manage the directions of knowledge transfer, so as to distill knowledge from both node-level and structure-level aspects.
In addition, we introduce FreeKD-Prompt to learn distorted and diverse graph augmentations for distilling varied knowledge. Furthermore, we develop FreeKD++ and FreeKD-Prompt++ to facilitate free-direction knowledge transfer among multiple shallow GNNs. Extensive experiments demonstrate the effectiveness of our methods. 

Our work primarily focuses on static graphs with a single type of relationship between nodes.
A future direction of our work is extending our approach to more complicated graphs such as heterogeneous graphs and dynamic graphs.
Furthermore, our method has been evaluated on the node classification task. In the future, we could extend our method to other graph tasks, such as link prediction and community detection, to explore its effectiveness in diverse graph learning scenarios.


%





\ifCLASSOPTIONcaptionsoff
  \newpage
\fi



%
\bibliographystyle{IEEEtran}
\bibliography{reference}


%

\begin{IEEEbiography}[{\includegraphics[width=1in,height=1.25in,clip,keepaspectratio]{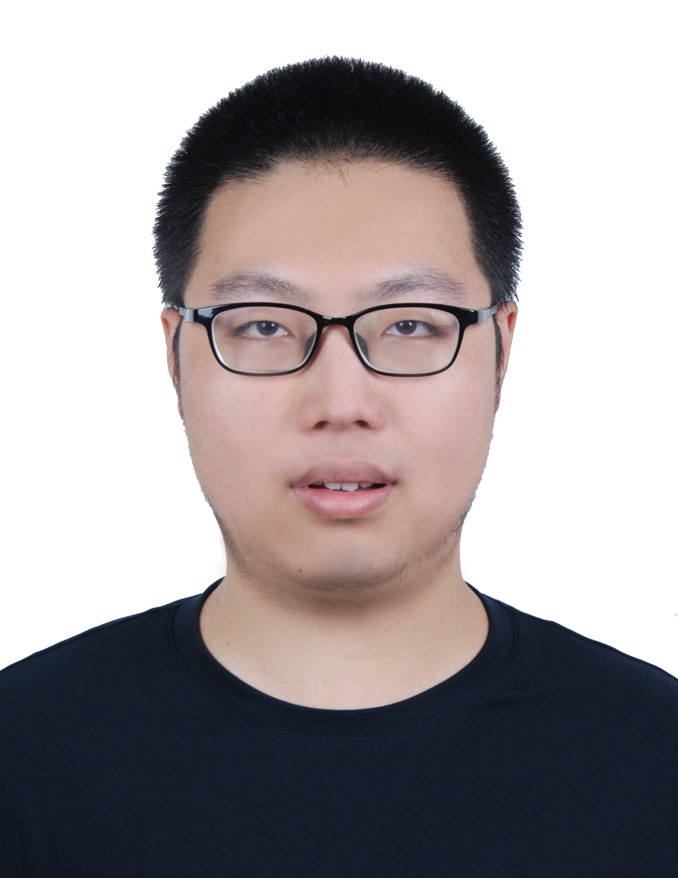}}]{Kaituo Feng}
received the B.E. degree in Computer Science and Technology from Beijing
Institute of Technology (BIT) in 2022. He is currently pursuing the master degree in Computer Science and Technology at Beijing Institute of Technology (BIT). His research interests include graph neural networks and knowledge distillation.
\end{IEEEbiography}
\vspace{-0.4in}
\begin{IEEEbiography}[{\includegraphics[width=1in,height=1.25in,clip,keepaspectratio]{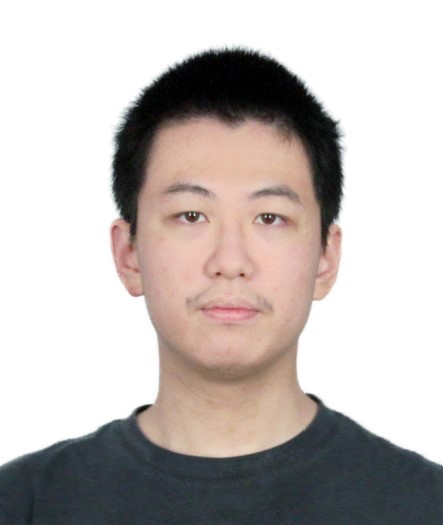}}]{Yikun Miao} is currently pursuing the B.E. degree in Computer Science and Technology at Beijing Institute of Technology (BIT). His research interests include machine learning and data mining.
\end{IEEEbiography}
\vspace{-0.4in}
\begin{IEEEbiography}[{\includegraphics[width=1in,height=1.25in,clip,keepaspectratio]{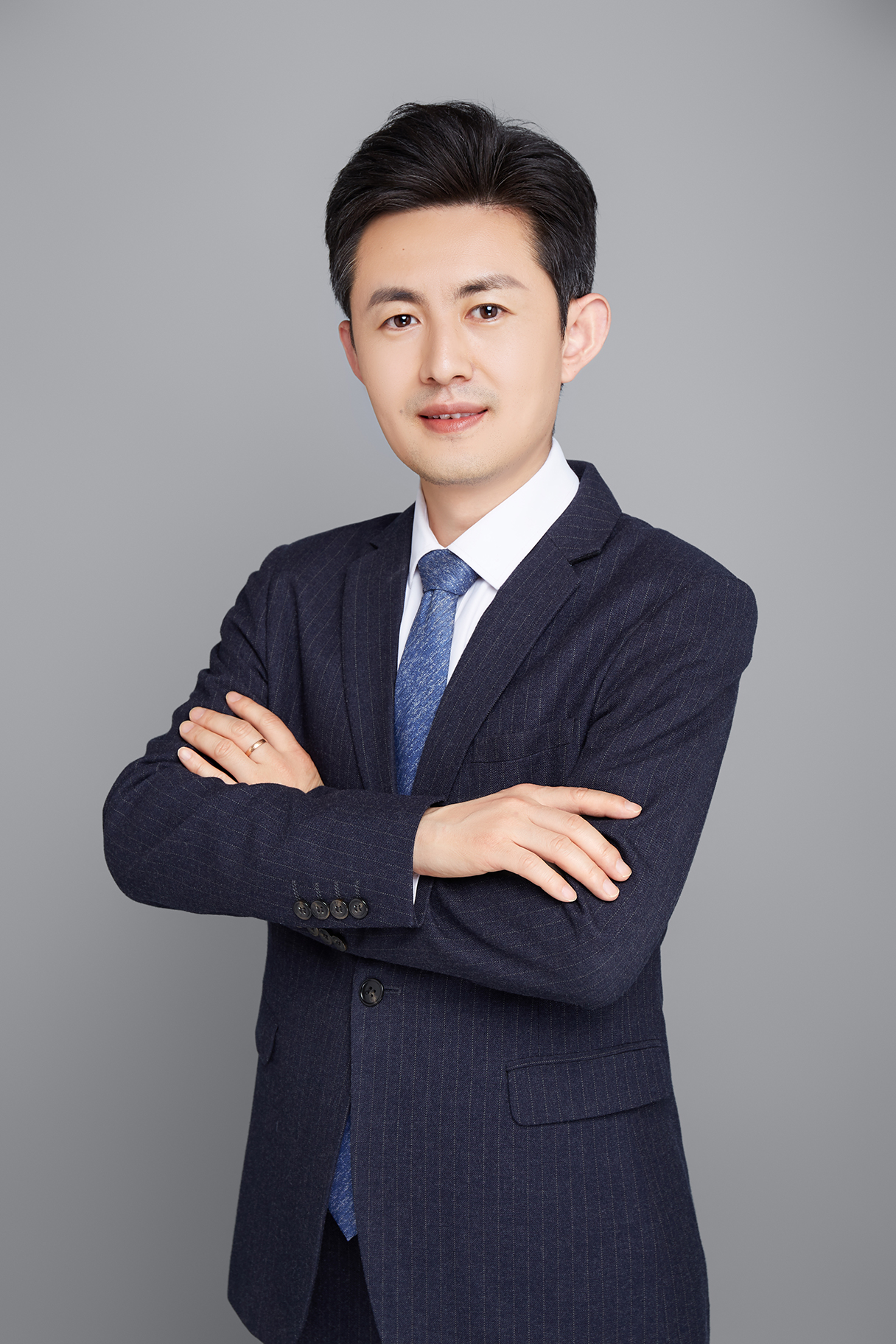}}]{Chengsheng Li}
received the B.E. degree from the University of Electronic Science and Technology of China (UESTC) in 2008 and the Ph.D. degree in pattern recognition and intelligent system from the Institute of Automation, Chinese Academy of Sciences, in 2013. During his Ph.D., he once studied as a Research Assistant with The Hong Kong Polytechnic University from 2009 to 2010. He is currently a Professor with the Beijing Institute of Technology. Before joining the Beijing Institute of Technology, he worked with IBM Research, China, Alibaba Group, and UESTC. He has more than 70 refereed publications in international journals and conferences, including IEEE TRANSACTIONS ON PATTERN ANALYSIS AND MACHINE INTELLIGENCE, IEEE TRANSACTIONS ON IMAGE PROCESSING, IEEE TRANSACTIONS ON NEURAL NETWORKS AND LEARNING SYSTEMS, IEEE TRANSACTIONS ON COMPUTERS, IEEE TRANSACTIONS ON MULTIMEDIA, PR, CVPR, AAAI, IJCAI, CIKM, MM, and ICMR. His research interests include machine learning, data mining, and computer vision. He won the National Science Fund for Excellent Young Scholars in 2021.
\end{IEEEbiography}
\vspace{-0.4in}
\begin{IEEEbiography}[{\includegraphics[width=1in,height=1.25in,clip,keepaspectratio]{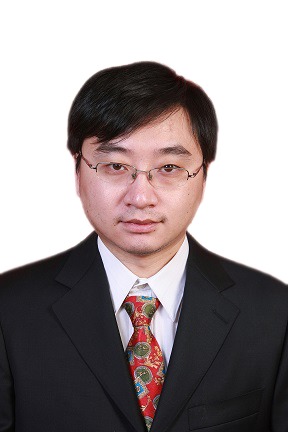}}]{Ye Yuan}
received the B.S., M.S., and Ph.D. degrees in computer science from Northeastern University in 2004, 2007, and 2011, respectively. He is currently a Professor with the Department of Computer Science, Beijing Institute of Technology, China. He has more than 100 refereed publications in international journals and conferences, including VLDBJ, IEEE TRANSACTIONS ON PARALLEL AND DISTRIBUTED SYSTEMS, IEEE TRANSACTIONS ON KNOWLEDGE AND DATA ENGINEERING,SIGMOD, PVLDB, ICDE, IJCAI, WWW, and KDD. His research interests include graph embedding, graph neural networks, and social network analysis. He won the National Science Fund for Excellent Young Scholars in 2016.
\end{IEEEbiography}
\vspace{-0.4in}
\begin{IEEEbiography}[{\includegraphics[width=1in,height=1.25in,clip,keepaspectratio]{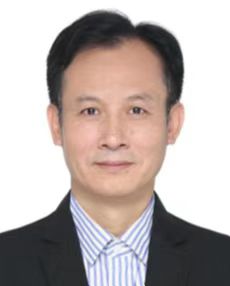}}]{Guoren Wang}
received the B.S., M.S., and Ph.D. degrees in computer science from Northeastern University, Shenyang, in 1988, 1991, and 1996, respectively. He is currently a Professor with the School of Computer Science and Technology, Beijing Institute of Technology, Beijing, where he has been the Dean since 2020. He has more than 300 refereed publications in international journals and conferences, including VLDBJ, IEEE TRANS-ACTIONS ON PARALLEL AND DISTRIBUTED SYSTEMS, IEEE TRANSACTIONS ON KNOWLEDGE AND DATA ENGINEERING, SIGMOD, PVLDB, ICDE, SIGIR, IJCAI, WWW, and KDD. His research interests include data mining, database, machine learning, especially on high-dimensional indexing, parallel database, and machine learning systems. He won the National Science Fund for Distinguished Young Scholars in 2010 and was appointed as the Changjiang Distinguished Professor in 2011.
\end{IEEEbiography}







\end{document}